\documentclass[10pt,journal,compsoc]{IEEEtran}

\usepackage{graphicx}
\usepackage{comment}
\usepackage{amsmath,amssymb} % define this before the line numbering.
\usepackage{color}
\newcommand{\tabincell}[2]{\begin{tabular}{@{}#1@{}}#2\end{tabular}}
\usepackage{makecell}
\usepackage{caption}
\usepackage{enumerate}
\usepackage{multirow}
\usepackage{multicol}
\usepackage{subfigure}
\usepackage{float}
\usepackage{bbding}

\usepackage{mathtools}
\usepackage[linesnumbered,ruled,vlined]{algorithm2e}
\usepackage{xcolor}

\SetCommentSty{mycommfont}

\begin{document}

\title{Global Instance Tracking: Locating Target More Like Humans}

\author{Shiyu~Hu,
        Xin~Zhao,~\IEEEmembership{Member,~IEEE,}
        Lianghua~Huang,
        and~Kaiqi~Huang,~\IEEEmembership{Senior~Member,~IEEE}

\IEEEcompsocitemizethanks{\IEEEcompsocthanksitem S. Hu is with the School of Artificial Intelligence, University of Chinese Academy of Sciences, Beijing 100049, China, and also with the Center for Research on Intelligent System and Engineering, Institute of Automation, Chinese Academy of Sciences, Beijing 100190, China.\protect
E-mail: hushiyu2019@ia.ac.cn.
\IEEEcompsocthanksitem X. Zhao is with the Center for Research on Intelligent System and Engineering, Institute of Automation, Chinese Academy of Sciences, Beijing 100190, China, and also with the University of Chinese Academy of Sciences, Beijing 100049, China.\protect
E-mail: xzhao@nlpr.ia.ac.cn.
\IEEEcompsocthanksitem L. Huang is with the Center for Research on Intelligent System and Engineering, Institute of Automation, Chinese Academy of Sciences, Beijing 100190, China.\protect
E-mail: huanglianghua2017@ia.ac.cn.
\IEEEcompsocthanksitem K. Huang is with the Center for Research on Intelligent System and Engineering and National Laboratory of Pattern Recognition, Institute of Automation, Chinese Academy of Sciences, Beijing 100190, China, and also with the University of Chinese Academy of Sciences, Beijing 100049, China, and the CAS Center for Excellence in Brain Science and Intelligence Technology, Beijing 100190, China.\protect
E-mail: kqhuang@nlpr.ia.ac.cn.}
\thanks{(Corresponding author: Xin Zhao)}}

\markboth{Journal of \LaTeX\ Class Files,~Vol.~, No.~, Month}%
{Shell \MakeLowercase{\textit{et al.}}: Bare Demo of IEEEtran.cls for Computer Society Journals}

\IEEEtitleabstractindextext{
\begin{abstract}
  Target tracking, the essential ability of the human visual system, has been simulated by computer vision tasks. However, existing trackers perform well in austere experimental environments but fail in challenges like occlusion and fast motion. The massive gap indicates that researches only measure tracking performance rather than intelligence. How to scientifically judge the intelligence level of trackers? Distinct from decision-making problems, lacking three requirements (a challenging task, a fair environment, and a scientific evaluation procedure) makes it strenuous to answer the question.
  In this article, we first propose the \textbf{global instance tracking (GIT)} task, which is supposed to search an arbitrary user-specified instance in a video without any assumptions about camera or motion consistency, to model the human visual tracking ability. Whereafter, we construct a high-quality and large-scale benchmark \textbf{VideoCube} to create a challenging environment. Finally, we design a scientific evaluation procedure using human capabilities as the baseline to judge tracking intelligence. Additionally, we provide an online platform with toolkit and an updated leaderboard. Although the experimental results indicate a definite gap between trackers and humans, we expect to take a step forward to generate authentic human-like trackers. The database, toolkit, evaluation server, and baseline results are available at http://videocube.aitestunion.com.
\end{abstract}

}

\maketitle

\IEEEraisesectionheading{\section{Introduction}\label{sec:introduction}}

\begin{figure*}[h!]
  \centering
  \includegraphics[width=0.95\textwidth]{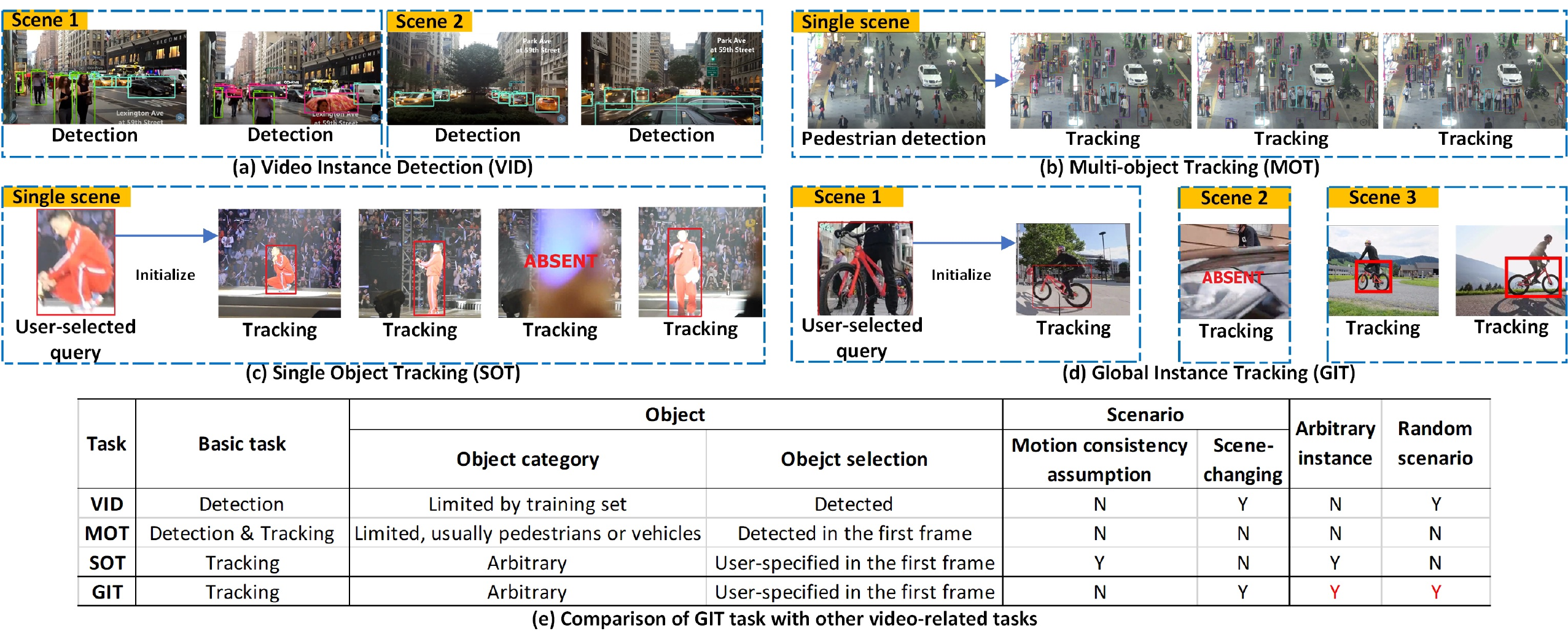}
  \caption{The execution flow and comparison of GIT with other video-related vision tasks (VID, MOT, SOT). VID (\textbf{a}) and MOT (\textbf{b}) can only locate limited instances, while SOT (\textbf{c}) and GIT (\textbf{d}) do not constrain the target category. Furthermore, GIT expands the SOT task by canceling the motion continuity assumption, allowing the target to move in a broader and more complex environment. The detailed comparison of GIT with above vision tasks is listed in \textbf{e}. Obviously, GIT is a new visual task without restrictions on target categories and scenarios.}
  \label{fig:task}
  \end{figure*}

\begin{figure}[h!]
  \centering
  \includegraphics[width=0.95\linewidth]{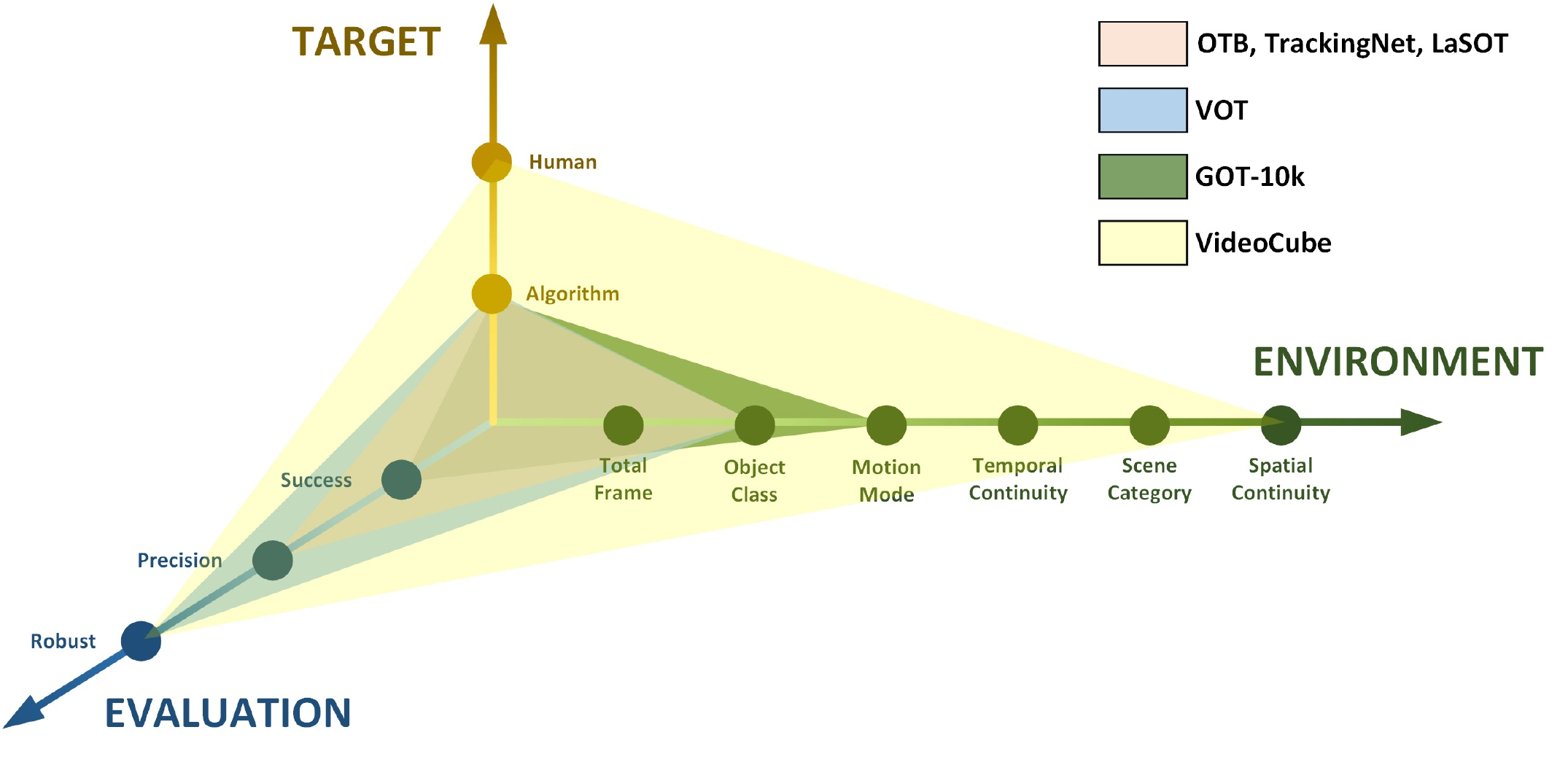}
  \caption{Comparison of VideoCube and other tracking benchmarks (OTB2015 \cite{OTB2015}, TrackingNet \cite{TrackingNet}, LaSOT \cite{LaSOT}, VOT2017 \cite{VOT2017}, GOT-10k \cite{GOT-10k}) in the complexity of the environment, the rationality of evaluation, and the completeness of target selection. }
  \label{fig:videocube1}
  \end{figure}

\begin{figure*}[h!]
  \centering
  \includegraphics[width=0.95\linewidth]{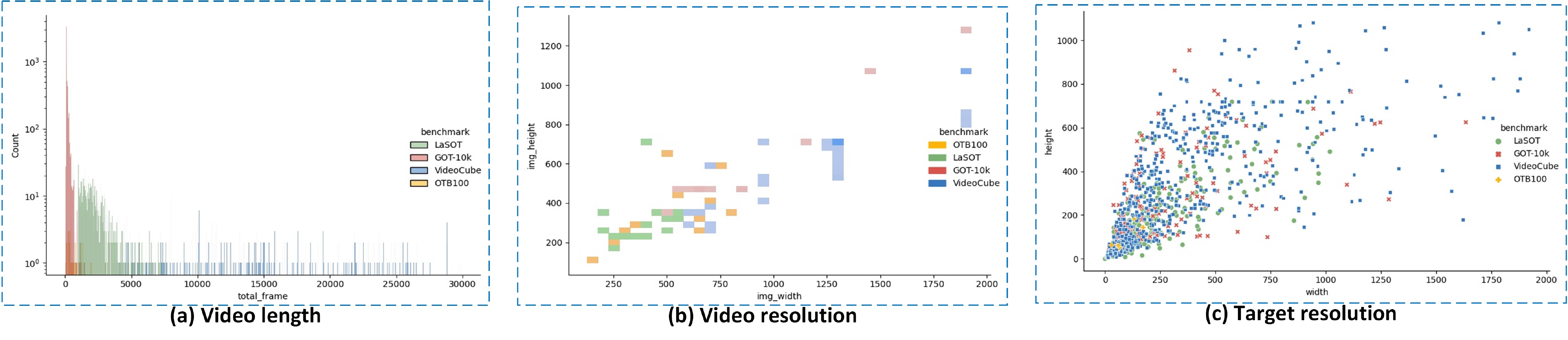}
  \caption{Comparison of VideoCube and three representative tracking benchmarks (OTB2015 \cite{OTB2015}, LaSOT \cite{LaSOT}, GOT-10k \cite{GOT-10k}) in video length (\textbf{a}), video resolution (\textbf{b}), and target resolution (\textbf{c}).}
  \label{fig:videocube2}
  \end{figure*}

\begin{figure*}[h!]
  \centering
  \includegraphics[width=0.95\linewidth]{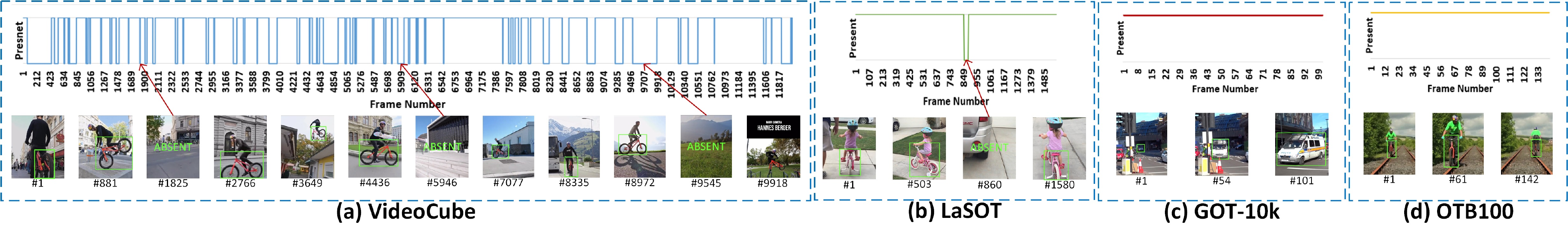}
  \caption{Comparison of VideoCube (\textbf{a}) and three representative tracking benchmarks (LaSOT \cite{LaSOT} (\textbf{b}), GOT-10k \cite{GOT-10k} (\textbf{c}), OTB2015 \cite{OTB2015} (\textbf{d})) in video content, video length, number of disappearances, and the absent duration.}
  \label{fig:videocube3}
  \end{figure*}

\begin{figure}[t!]
  \centering
  \includegraphics[width=0.95\linewidth]{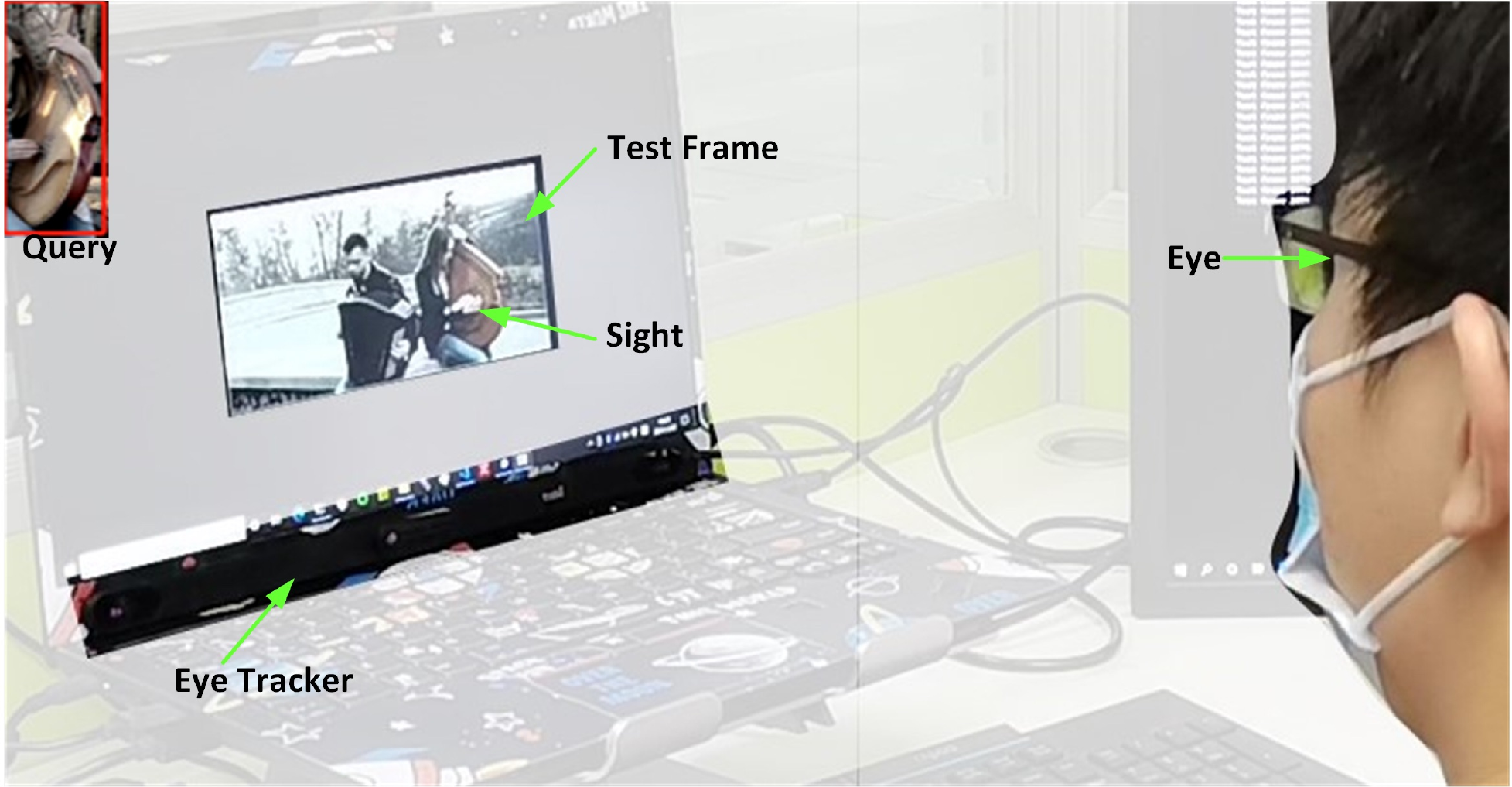}
  \caption{Schematic diagram of the human visual tracking experiment.}
  \label{fig:eye}
  \end{figure}

\IEEEPARstart{T}{arget} tracking, the ability to follow a moving object with the human eyes, is the basic function of the human visual system. 
Research reveals that a baby can master this skill at only a few weeks of age and quickly expand from tracking salient objects (e.g., a brightly colored toy) to arbitrary objects (e.g., a decoration on the clothes of parents) \cite{baby1,baby2}. 
Inspired by the powerful human visual system and eye-catching artificial intelligence technology, researchers have proposed a series of visual tasks to locate moving targets in the real environment. 
Several existing computer vision tasks, such as single object tracking (SOT \cite{SOT1}), multi-object tracking (MOT \cite{MOT1}), and visual instance detection (VID \cite{VOD2}), simulates human target tracking ability to locate moving targets in the natural environment, and are widely used in animal behavior observation \cite{dankert2009automated, weissbrod2013automated,mathis2018deeplabcut,wei2018behavioral}, medical research \cite{celltracking,namboodiri2019single,payton2007biomechanical} and robot navigation \cite{robot-navigation}. 

However, challenging conditions like occlusion, fast motion, and weak illumination reduces the performance of existing methods. Take automatic driving as an example - several crashes happened at night or under bright light conditions due to the limit in visual perception robustness of trackers, which contrasts to the high performance judged by the vision task benchmarks. In other words, existing experimental environments only measure performance rather than intelligence, far away from the actual applications. A natural question is, how to scientifically measure the tracking intelligence of an algorithm?

The imitation game proposed by Alan Turing in 1950 \cite{turing}, which is usually called the Turing test, is a recognized standard to judge machine intelligence. 
Recently, the agents represented by AlphaGo (Go game \cite{silver2017mastering} AI) and DeepStack (Poker game \cite{brown2018superhuman} AI) have defeated the top human professional players in decision-making problems, and become the landmark results of the Turing Test. From these works, we can summarize three requirements for machine intelligence measurement: (1) a challenging task (e.g., Go game is difficult for both humans and machines); (2) a fair competition environment (e.g., human and machine compete in the Go game with equal rules); and (3) a scientific evaluation procedure (e.g., players with a larger number of vacant intersections and captured stones win the Go game). 
Nevertheless, the existing target tracking area lacks these three points, making it strenuous to evaluate visual intelligence. 

For the first requirement, a proper task is essential to estimate visual tracking intelligence. Simple assignments (such as tracking a black dot on a white screen) cannot reflect intelligence, while unmanageable tasks (such as tracking an ant in a colony with a shaking camera) are almost impossible for humans to execute. Therefore, the reasonable idea is to design a moderately difficult task based on human visual tracking ability. Clearly, people can unconsciously locate an \textit{arbitrary instance} in \textit{random scenarios}, while the existing tasks always contain strong constraints on target categories (MOT, VID) or scenarios (SOT).
% For example, humans can quickly locate and follow an arbitrary actor in the multi-viewed videos, but the existing tasks cannot directly describe this scene.

As the second requirement, a suitable benchmark needs to reflect the characteristics of the task and simulate the natural environment. Dynamic visual acuity, the essential human ability to perceive moving objects, can be improved by tracking fast-moving targets in complex environments.
Thus, a decent benchmark should fitly reproduce the proximate real-world conditions and provide a platform for training a human-like tracker.
However, existing tracking benchmarks only provide a simplistic environment. Trackers generated by these benchmarks are still far from the human visual system and cannot suit challenging realistic conditions like occlusion, fast motion, and weak illumination.

The last requirement, a scientific evaluation system, should set targets (machine and human) into the same environment and measure their tracking capabilities with reasonable indicators. 
Unlike Go and poker games with clear rules, trackers and humans have exceptionally distinct ways of performing visual tracking tasks. Algorithms usually process the video frame by frame and return bounding boxes to locate the object, while humans directly focus their sight on the target. 
Existing benchmarks are all designed for evaluating algorithms but lack standards for measuring human tracking ability. Lacking the comparison with humans means we cannot measure the intelligence level of algorithms accurately.

Based on the above three problems, this work evaluates tracking intelligence degree for the first time by providing: 

\noindent
\textbf{(1) A proper task to model human visual tracking ability.}
We introduce \textbf{global instance tracking (GIT)}, a new task of searching an arbitrary user-specified instance in a video without any assumptions on camera or motion consistency, to accurately model the human tracking ability.
Unlike the existing video-related tasks, GIT aims to find all video fragments where a query object presents and locates its trajectories in these fragments. GIT retains the category-independent advantage and expands the boundary of the traditional SOT task to approach object tracking in general scenes. An ideal GIT algorithm is supposed to work in different video environments like rapid view angle changes, frequent camera switches, or long-term target absences.
The execution flow and comparison of GIT with other video-related vision tasks are shown in Figure~\ref{fig:task}.

\noindent
\textbf{(2) A comprehensive benchmark to simulate the real world.}
We provide a high-quality, large-scale benchmark \textbf{VideoCube} for this novel task. It consists of 500 long-term videos that cover different object classes, scenario types, motion modes, and challenge attributes, with an average length of \textit{14920} frames. Figure~\ref{fig:videocube1} to Figure~\ref{fig:videocube3} illustrates that by comparing with existing visual tracking benchmarks, VideoCube provides a proximate real-world environment and evaluates the algorithms scientifically.

\noindent
\textbf{(3) A scientific evaluation procedure to compare humans and machines with reasonable indicators.}
In addition to evaluating trackers via classical metrics, we judge \textbf{human visual tracking capability} via an eye-tracking experiment for the first time. Figure~\ref{fig:eye} is the schematic diagram of the human visual tracking experiment. Human performance is treated as a baseline to measure the intelligence level of existing methods. 
The result illustrates that SOTA trackers can perform well in a simple situation (target with smooth movement) but fail in difficulties (e.g., occlusion, fast motion, and weak illumination), while humans can still maintain fast and accurate tracking with challenging factors.

Besides, we provide a comprehensive online platform at http://videocube.aitestunion.com with systematic evaluation toolkits, an online evaluation server, and a real-time leaderboard. We believe the online platform with the human baseline can provide researchers with more comprehensive assistance in visual intelligence research and take a step forward to generate authentic human-like trackers.

The rest of this paper is organized as follows. Section~\ref{sec:related_work} provides a review of video-related tasks and distinguishes them from GIT. Section~\ref{sec:videocube} introduces the design principles of VideoCube. The experimental results and detailed analysis are described in Section~\ref{sec:experiments}. Finally, we conclude this paper and discuss future works in Section~\ref{sec:conclusion}.

\section{Related Work}
\label{sec:related_work}

Capturing local motion and predicting long-term moving trajectories of targets in a video is of great significance to many research fields \cite{idtracker,celltracking}. Several vision tasks have been modeled for locating moving objects in video.  This section introduces these visual tasks' definitions, characteristics, and application scenarios to distinguish them from GIT. 

\subsection{Locate specific target categories in random scenarios}

Video instance detection (VID) \cite{VOD2} is a fundamental prerequisite for advanced visual tasks such as scene content analysis and understanding. It aims to accurately determine the category and location of each target in a video. The target category is generally limited to the known classes in the training dataset, but the video without any restrictions may contain various scenes.

Multiple object tracking (MOT) \cite{MOT1} is a model-specific visual task that focuses on tracking specific categories like persons or vehicles without any prior knowledge about the appearance and amount. The general MOT algorithm usually runs a detector to obtain the object's bounding box in the first frame and generate features; then calculates the similarity to determine instances belonging to the same target and assigns a digital ID to each object. 

\subsection{Track random objects in a single scenario}

Single object tracking (SOT) \cite{SOT1} intends to calculate the location of a user-specific visual target in the video when only a position in the first frame is available. Unlike other visual tasks, SOT is an entirely category-independent assignment suitable for open-set testing with broad prospects. However, the implicit motion continuity assumption limits its actual applications. Since SOT is the vision task closest to GIT in assignment settings, the following part introduces the related trackers and benchmarks in detail.

\subsubsection{Trackers}

%Visual object tracking has accomplished dramatic advancement in the past few decades, including improving accuracy, enhancing robustness, and accelerating speed. Here we divide these methods into correlation-filter trackers and deep trackers to arrange a brief review.

\noindent
\textbf{Correlation-filter trackers}.
Correlation-filter (CF) trackers regard the SOT task as a regression problem and achieve high speed via fast Fourier transform (FFT) \cite{KCF}. Dense image sampling by circulant shift on a single centered image patch is essential to implement fast training and inference in the Fourier domain. 
As the first model to utilize the correlation filter framework in object tracking, MOSSE \cite{MOOSE} considers this task as a regularized least-squares problem and reformulates its closed-form solution, achieving reliable tracking performance at 700 fps.
Later on, several improvements have been proposed, including using a scale embedding to handle scale variation \cite{danelljan2014accurate} and improving CF tracking via extra regularization method \cite{danelljan2015learning}. 

\noindent
\textbf{Deep trackers}. 
Recently, several methods based on deep learning have been proposed to advance tracking performance. 
Convolutional neural networks (CNNs) are the most widely-used model, involving extracting features through pre-trained models \cite{ECO} or using end-to-end learning to generate object appearance models \cite{SiamFC}. 
The siamese trackers \cite{SiamFC,tao2016siamese} and their variants \cite{SiamRPN,galoogahi2017learning} regard tracking as a feature matching task and achieve a significant result. 
By learning a high dimensional metric space between the exemplar and search patches, siamese trackers can quickly localize the instance in a consecutive sequence. 
Except for CNN-based models, some advanced deep trackers regard tracking as a sequential decision-making task \cite{yun2017action}, or combine the recurrent structures to accomplish sequential prediction \cite{yang2017recurrent}. 

\subsubsection{Benchmarks}

%A high-quality benchmark labels the target in the video frame and provides criteria for algorithm evaluation, which plays a vital role in SOT development. The early benchmarks represented by OTB \cite{OTB2013,OTB2015} are mainly designed for short-term tracking tasks and consist of short videos without target-absent labels \cite{VOT2017,TC128,UAV2016,NUS-PRO2015,Nfs2017}, which assumes no complete occlusion or target out-of-view happened in this video. In recent years, the development of deep learning boosts the demand for large-scale and high-quality benchmarks \cite{TrackingNet,OxUvA,GOT-10k,LaSOT}. SOT benchmarks have developed in two directions: the first represented by GOT-10k \cite{GOT-10k} aims to improve generalization of short-term trackers by including enough object categories and using zero-overlapped strategy in splitting dataset. LaSOT \cite{LaSOT} stands for the second direction, which allows a transitory target disappearance-reappearance process and expands the task to the long-term scenarios.

\noindent
\textbf{Short-term tracking benchmarks}.
A series of benchmarks have appeared since 2013 and provide a consolidated platform for evaluating and analyzing algorithms. As one of the earliest benchmarks, OTB2013 \cite{OTB2013} includes 51 fully-labeled short sequences and evaluates the performance of the previous 29 top trackers. Subsequently, OTB2015 \cite{OTB2015} expands the benchmark to 100 videos to provide unbiased performance comparisons. The VOT \cite{VOT2013, VOT2014,VOT2015,VOT2016,VOT2017,VOT2018,VOT2019} has been an annual visual object tracking challenge since 2013, which provides a diverse and adequately small dataset from existing visual tracking datasets. TC-128 \cite{TC128} collects and annotates 78 new videos based on OTB2013 \cite{OTB2013} to provide the evaluation of color-enhanced tracking algorithms on color sequences. Several datasets are designed for tracking specific instances. The NUS-PRO \cite{NUS-PRO2015} dataset focuses on tracking pedestrian and rigid objects, and the UAV123 \cite{UAV2016} comprises 123 short videos for assessing unmanned aerial vehicle tracking performance. Nfs \cite{Nfs2017} provides 100 sequences with a higher frame rate (240 FPS) camera, intending to examine the trade-off bandwidth limitations related to real-time analysis of visual trackers.
With the advancement of deep learning, a large-scale and high-quality dataset for short-term tracking is demanded. GOT-10k \cite{GOT-10k} is a significant high-diversity benchmark and comprises 10,000 videos from the semantic hierarchy of WordNet \cite{WordNet} to accommodate plentiful object categories and motion trajectories. It is the first benchmark to suggest the one-shot protocol for evaluating tracking performance and improving model generalization.

\noindent
\textbf{Long-term tracking benchmarks}.
Allowing brief disappearance and having a longer duration are two characteristics of long-term tracking. OxUvA \cite{OxUvA} is the first large-scale dataset for this task and selects 366 videos with an average duration of 144 seconds, but only performs annotation every 30 frames. LaSOT \cite{LaSOT} is first released in 2019 and provides a dataset with 3.5M manually labeled frames, including 1400 videos with 70 categories. In 2020, LaSOT is expanded to 1550 videos and 85 classes. It is re-divided with the one-shot protocol of GOT-10k \cite{GOT-10k} to improve the generalization.

Consequently, SOT can continuously locate objects of any category due to model-free characteristics and is more versatile for open-set test environments. However, the existing SOT methods are still far from robust long-term tracking in complex environments for three reasons:(1) Strong constraints in the task definition. The implicit continuous motion assumption limits the task environment in continuous-time and single-scene, far from the natural application environment. (2) Limited video type in the existing benchmarks. Videos with a single shot and a single scene cannot fully reflect the complexity of the actual situations. (3) Strong timing-dependence in the modeling process, which accumulates errors and cannot achieve robust tracking in long-term tracking.

\begin{table*}[h!]
  \begin{center}
  \caption{\textbf{Comparison of VideoCube with popular single object tracking benchmarks.} VideoCube is superior to existing datasets in multiple dimensions, including scale, label density, and content richness (object classes, motion modes, scene categories). 
  Note: (a) TrackingNet performs manual annotation per second and uses the DCF \cite{DCF} algorithm to automatically label the remaining frames to accomplish dense labeling with 30Hz frequency. 
  (b) GOT-10k extracts 1.45 million images from more than 40h videos at 10FPS and manually annotates each frame. 
  (c) The object classes in GOT-10k are finely divided based on WordNet \cite{WordNet}. For example, the border collie is an independent category, rather than being divided into dogs. 
  (d) OxUvA believes that the manual labeling frequency of 1 Hz is sufficient for trackers, thus only offering annotation once per second. 
  (e) OxUvA only performs additional annotation about target absence but ignores other challenging attributes. 
  (f) VideoCube combines manual and automatic annotation similar to TrackingNet but increases the manual label frequency to 10Hz due to frequent scene switching in videos, and uses PrDiMP \cite{PrDiMP} to complete 30Hz dense annotation. 
  (g) VideoCube uses WordNet as the semantic framework to divide the video objects into 9 categories and 89 sub-categories. 
  (h) Given WordNet's limited ability to classify unique scenes, VideoCube uses WordNet as the backbone and references FrameNet \cite{FrameNet} and ConceptNet \cite{ConceptNet} to divide scenes into 8 categories and 55 sub-categories.}

  \scriptsize
  \begin{tabular}{p{1.8cm}<{\centering}|p{0.4cm}<{\centering}p{0.6cm}<{\centering}p{0.6cm}<{\centering}p{0.6cm}<{\centering}p{0.7cm}<{\centering}p{0.6cm}<{\centering}p{0.7cm}<{\centering}p{0.7cm}<{\centering}p{1.3cm}<{\centering}p{1cm}<{\centering}p{0.7cm}<{\centering}p{0.7cm}<{\centering}p{1cm}<{\centering}}
  \hline
  Benchmark & Year  & Videos & \tabincell{c}{Min \\ Frame} & \tabincell{c}{Mean \\ Frame} & \tabincell{c}{Median \\ Frame} & \tabincell{c}{Max \\ Frame} & \tabincell{c}{Total \\ Frame} & \tabincell{c}{Total \\ Duration} & \tabincell{c}{Label \\ Density} & \tabincell{c}{Attribute \\ Classes \\ (Absent)} & \tabincell{c}{Object \\ Classes} & \tabincell{c}{Motion \\ Modes}& \tabincell{c}{Scene \\ Categories} \\  
  \hline
  \textbf{OTB2013} \cite{OTB2013} & 2013 & 51 & 71 & 578 & 392 & 3872 & 29K & 16.4m & 30Hz & 11(\XSolidBrush) & 10 & n/a & n/a\\
  \textbf{OTB2015} \cite{OTB2015} & 2015 & 100 & 71 & 590 & 393 & 3872 & 59K & 32.8m & 30Hz & 11(\XSolidBrush) & 16 & n/a & n/a\\
  \textbf{TC-128} \cite{TC128} & 2015 & 129 & 71 & 429 & 365 & 3872 & 55K & 30.7m & 30Hz & 11(\XSolidBrush) & 27 & n/a & n/a\\
  \textbf{NUS-PRO} \cite{NUS-PRO2015} & 2015 & 365 & 146 & 371 & 300 & 5040 & 135K & 75.2m & 30Hz & n/a & 8 & n/a & n/a\\
  \textbf{UAV123} \cite{UAV2016} & 2016 & 123 & 109 & 915 & 882 & 3085 & 113K & 75.2m & 30Hz & 12(\XSolidBrush) & 9 & n/a & n/a\\
  \textbf{VOT-2017} \cite{VOT2017} & 2017 & 60 & 41 & 356 & 293 & 1500 & 21K & 11.9m & 30Hz & n/a & 24 & n/a & n/a\\
  \textbf{Nfs} \cite{Nfs2017} & 2017 & 100 & 169 & 3830 & 2448 & 20665 & 383K & 26.6m & 240Hz & 9(\XSolidBrush) & 17 & n/a & n/a\\
  \textbf{TrackingNet} \cite{TrackingNet} & 2018 & 30643 & - & 498 & -  & - & 14M & 141h & 1Hz(30Hz)$^{a}$ & 15(\XSolidBrush) & 27 & n/a & n/a\\
  \textbf{GOT-10k} \cite{GOT-10k} & 2019 & 10000 & 29 & 149 & 101 & 1418 & 1.45M & 40h & 10Hz$^{b}$ & 6(\checkmark) & 563$^{c}$ & 87 & n/a\\
  \hline
  \textbf{UAV20L} \cite{UAV2016} & 2016 & 20 & 1717 & 2934 & 2626 & 5527 & 59K & 32.6m & 30Hz & 12(\XSolidBrush) & 5 & n/a & n/a\\
  \textbf{OxUvA} \cite{OxUvA} & 2018 & 366 & 900 & 4320 & 2628 & 37740 & 1.55M & 14.4h &1Hz$^{d}$ & (\checkmark)$^{e}$ & 22 & n/a & n/a\\
  \textbf{LaSOT} \cite{LaSOT} & 2020 & 1550 & 1000 & 2502 & 2145 & 11397 & 3.87M & 35.8h & 30Hz & 14(\checkmark) & 85 & n/a & n/a\\
  \hline
  {\textbf{VideoCube}} & 2022 & 500 & \textcolor{red}{\textbf{4008}} & \textcolor{red}{\textbf{14920}} & \textcolor{red}{\textbf{14162}} & \textcolor{red}{\textbf{29834}} & \textcolor{red}{\textbf{7.46M}} & \textcolor{red}{\textbf{69.1h}} & 10Hz(30Hz)$^{f}$ & 12(\checkmark) & \textcolor{red}{\textbf{9(89)}}$^{g}$ & \textcolor{red}{\textbf{61}} & \textcolor{red}{\textbf{8(55)}}$^{h}$\\
  \hline
  \end{tabular}
  \label{table:benchmarks}
  \end{center}
  \end{table*}

\begin{figure}[h!]
  \centering
  \includegraphics[width=0.9\linewidth]{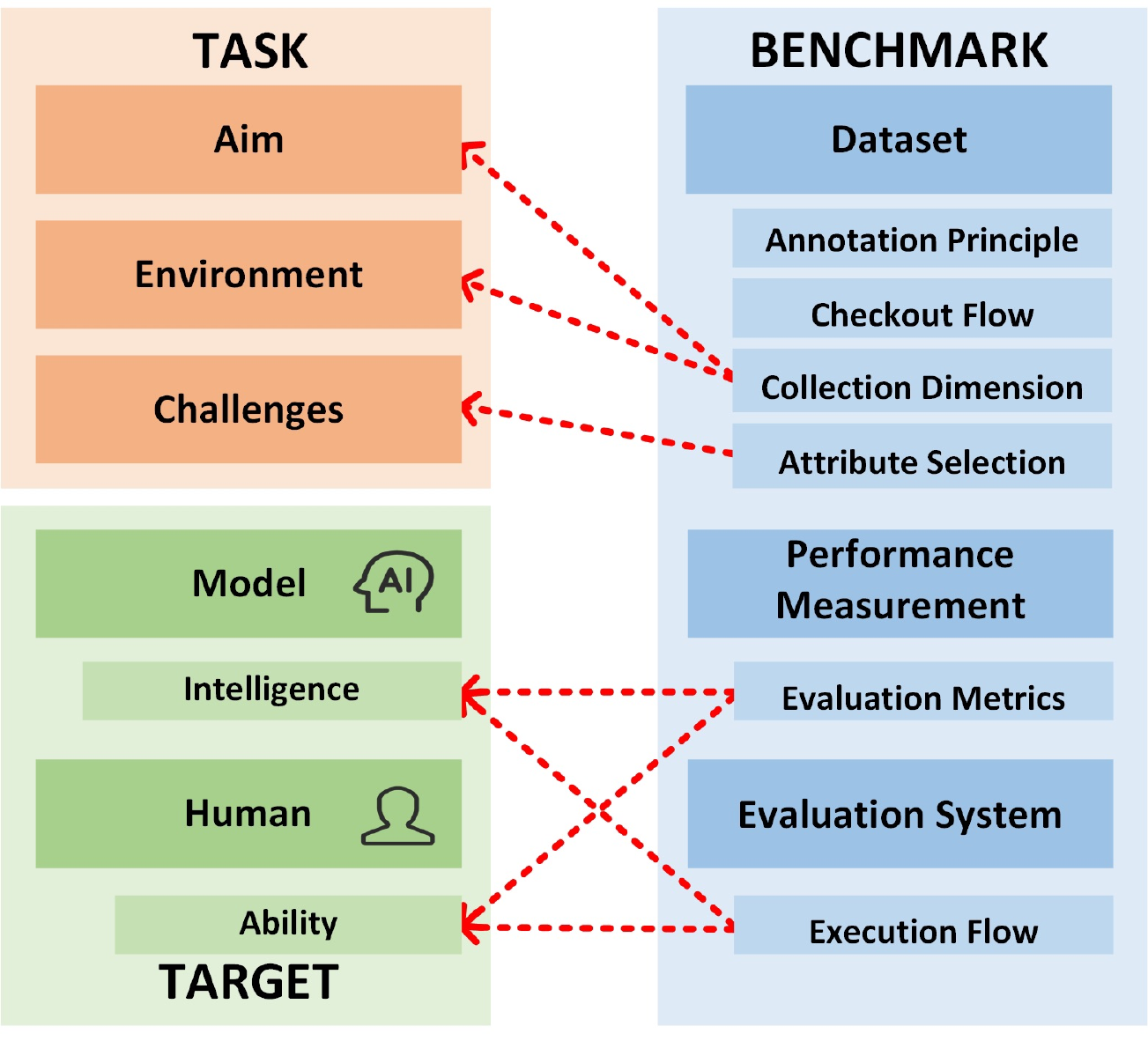}
  \caption{Construction principles of the VideoCube benchmark. We assume that a scientific benchmark should characterize the specified task and evaluate the model intelligence. Dataset, evaluation system, and performance measurement are three critical points included in constructing a benchmark. The red dotted line expresses the relationship of various fields.}
  \label{fig:3D}
  \end{figure}

\begin{figure*}[h!]
  \centering
  \includegraphics[width=0.9\textwidth]{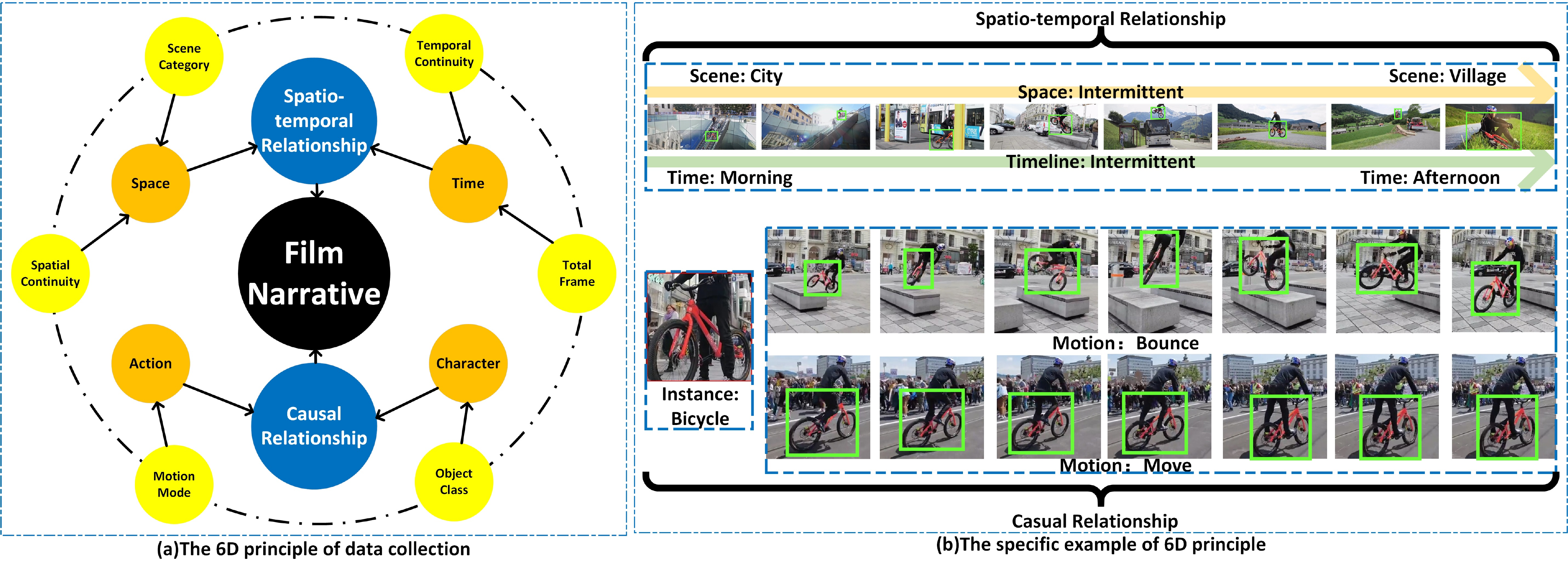}
  \caption{The 6D principle of data collection. We split the film's narrative into the spatio-temporal and causal relationship and further decompose them into six dimensions (scene category, spatial continuity, temporal continuity, total frame, motion mode, object class) to provide a more comprehensive description.}
  \label{fig:6D}
  \end{figure*}

\section{Construction of VideoCube}
\label{sec:videocube}

As a high-quality benchmark, VideoCube contains a large-scale dataset, reasonable evaluation metrics, and scientific evaluation systems to provide a general platform for intelligence measurement (Figure~\ref{fig:3D}).

\begin{figure*}[h!]
  \centering
  \includegraphics[width=0.95\textwidth]{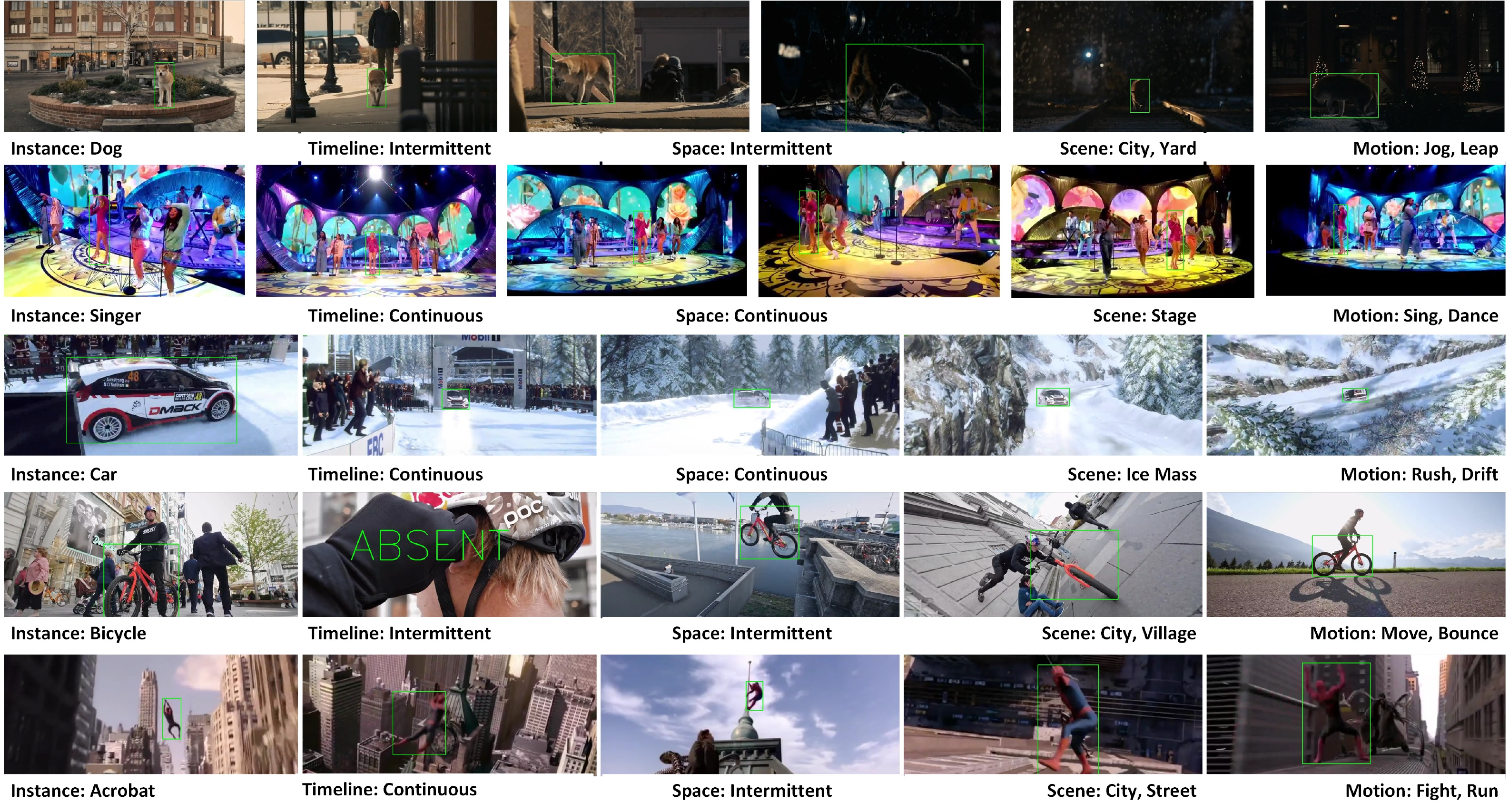}
  \caption{The representative data of VideoCube. Each video is strictly selected based on duration, instance classes, main scene categories, main motion modes, spatial consistency, and time consistency.}
  \label{fig:example}
  \end{figure*}

\begin{figure*}[h!]
  \centering
  \includegraphics[width=0.95\textwidth]{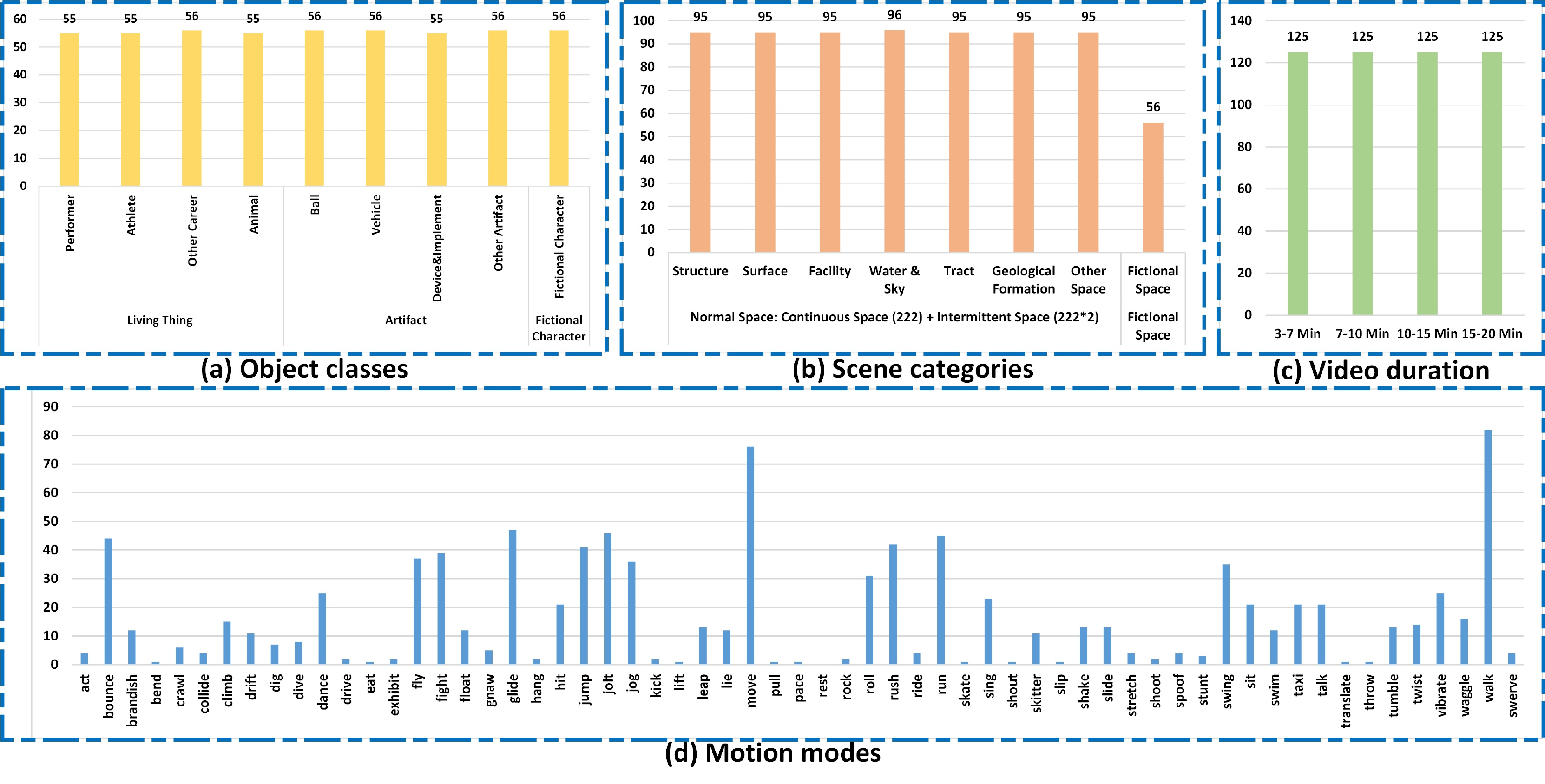}
  \caption{Data distribution of VideoCube.
  (\textbf{a}) The distribution of object classes.
  (\textbf{b}) The distribution of scene categories.
  (\textbf{c}) The distribution of video duration.
  (\textbf{d}) The distribution of motion modes.}
  \label{fig:distribution}
  \end{figure*}

\subsection{Dataset}

\textbf{VideoCube} is a reliable global instance tracking benchmark that contains scenes and instances adequately to reflect the diversity of real life. Before constructing it, we first summarize the key elements (e.g., benchmark, task, and target) and propose our design principles based on Figure~\ref{fig:3D}. Several aspects are considered in constructing this dataset:

\noindent
\textbf{(1) Multiple collection dimension}. 
The collection of VideoCube is based on six dimensions (Figure~\ref{fig:6D}) to describe the spatio-temporal relationship and causal relationship of film narrative, which provides an extensive dataset for the novel GIT task. We guarantee that each video contains at least \textit{4008} frames, and the average frame length in VideoCube is around \textit{14920}. Besides, the selected videos contain \textit{transitions} and target \textit{disappearance-reappearance} process to cancel the motion continuity assumption.

\noindent
\textbf{(2) Specific annotation principle and exhaustive checkout flow}.
A professional labeling team manually marked each video with a 10Hz annotation frequency, and all videos have passed three rounds of review by trained verifiers. Based on rigorous experiments, we selected the most effective algorithm PrDiMP \cite{PrDiMP} to combine manual annotations and accomplish intensive labels with 30Hz frequency.

\noindent
\textbf{(3) Comprehensive attribute selection}.
Multiple shots and frequent scene-switching make the video content change dramatically and become more challenging for algorithms. Thus, we accommodate twelve attributes annotations for each frame to implement a more elaborate reference for the performance analysis.

\subsubsection{Collection dimension}
 
The collection dimension is an essential basis for constructing datasets. Rich dimensions can restore the narrative content and simulate real application scenarios through dimensions integration. However, most existing video datasets only consider instance category and video duration when constructing but lack an overall narration expression.
As the scale of datasets has increased in recent years, several datasets have begun to extend their collection dimensions. For example, GOT-10k \cite{GOT-10k} combines the motion modes, and LaSOT \cite{LaSOT} adds a natural language description to characterize the video content. Nevertheless, we consider that the existing datasets lack a widespread meditation on dimension selection. 
The organization of instance categories and motion modes such as GOT-10k \cite{GOT-10k} is suitable for short-term rather than long-term tasks. The natural language description used by LaSOT \cite{LaSOT} seems to express the video content intuitively, but this annotation is subjective since personal views will inevitably be involved. Besides, an extra algorithm is needed to extract useful information in sentences, which increases the complexity of usage and errors.

How to determine the collection dimensions? The film narrative is defined as a chain of causal relationship events occurring in space and time \cite{Bordwell2011Film}. The causal relationship is determined by characters and activities, while the spatio-temporal relationship combines scene, time, and their continuity. Consequently, we connect scene category, spatial continuity, temporal continuity, total frame, motion mode, and object class as \textit{6D principle} (Figure~\ref{fig:6D}) to collect videos in VideoCube. The detailed introduction of 6D principle is organized as follows:

\begin{figure}[h!]
  \centering
  \includegraphics[width=0.9\linewidth]{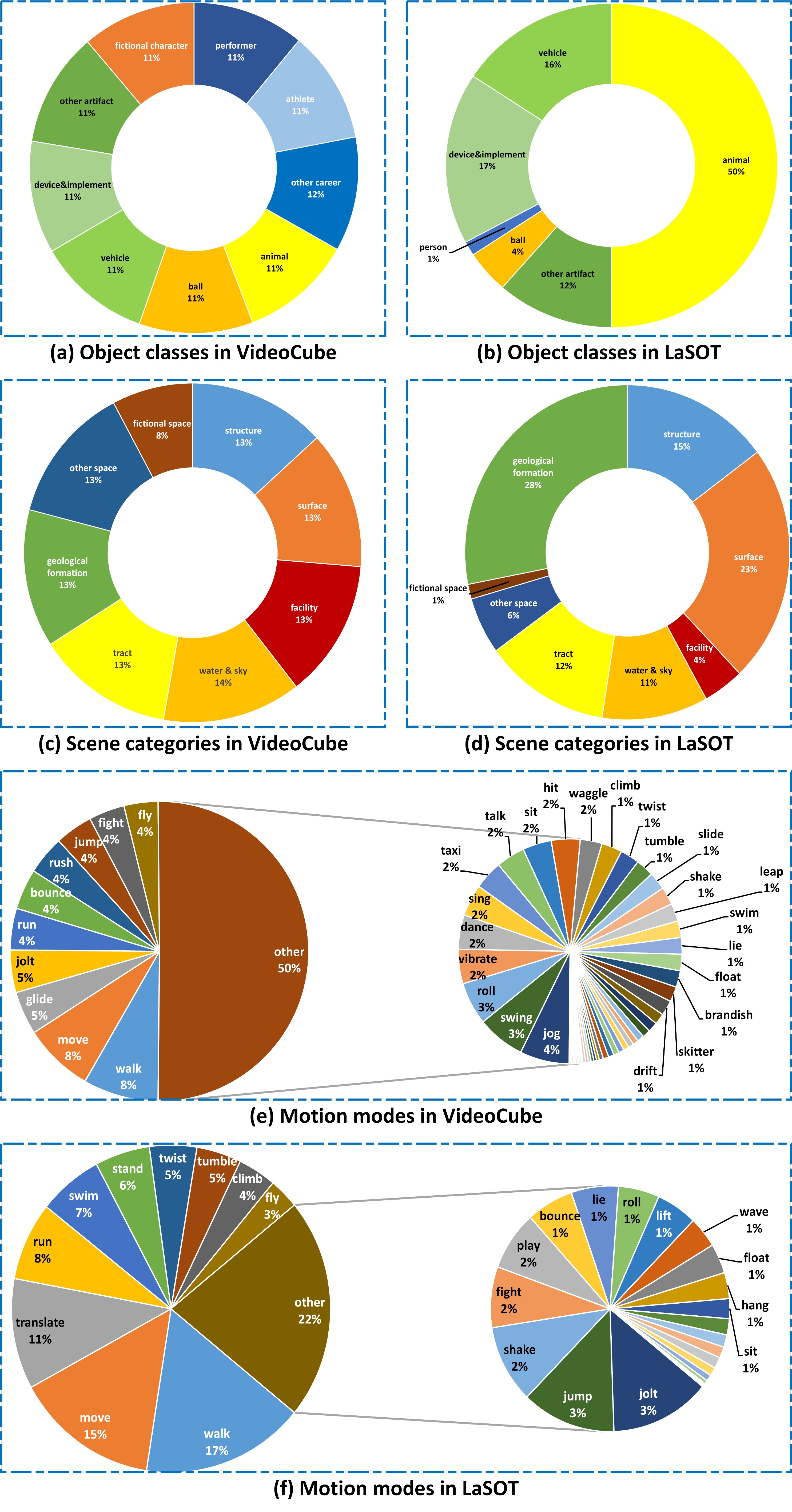}
  \caption{The distribution of object classes (\textbf{a-b}), scene categories (\textbf{c-d}) and motion modes (\textbf{e-f}) in VideoCube and LaSOT \cite{LaSOT} (based on WordNet \cite{WordNet}).}
  \label{fig:benchmark-compare}
  \end{figure}

\noindent
\textbf{Object classes}. 
Different from the existing datasets, VideoCube collects 89 typical instances and divides them into nine main categories based on the semantic framework WordNet \cite{WordNet}. As shown in Figure~\ref{fig:distribution} (a), it maintains an even distribution across the main categories. Since \emph{person} is the most common instance category while people with different identities have significant differences in motion modes and appearances, we split the person class into performer, athlete, and other careers.
Besides, given that computer-generated instances are common in some application scenarios but ignored by other datasets, we also add the functional character.

As shown in Figure~\ref{fig:benchmark-compare} (a-b), VideoCube has advantages in the distribution of object classes, and the nine root categories maintain uniform distribution. 
Although LaSOT \cite{LaSOT} maintains an even distribution on 70 classes, half of the data belong to the \emph{animal} category, while only 20 sequences (1.43\%) belong to the \emph{person}.

\noindent
\textbf{Spatial continuity and scene categories}. 
Videos in VideoCube are divided into normal space and fictional space. First of all, 56 videos (to keep the same video amount with the fictional character in Figure~\ref{fig:distribution} (a)) are reserved as the fictional space. Since VideoCube cancels the motion continuity assumption, the instance may occur in multiple scenes, causing scene-switching in a video. Therefore, we divide the normal space videos into 222 continuous spaces and 222 intermittent spaces, then record the single scene of continuous space and two main scenes of intermittent space. Finally, all the 666 scenes are evenly divided into seven main categories, as shown in Figure~\ref{fig:distribution} (b).

Figure~\ref{fig:benchmark-compare} (c-d) exhibits the distribution of scene categories in VideoCube and LaSOT \cite{LaSOT}. Since the object class of LaSOT \cite{LaSOT} is mainly animals, its scene categories are primarily concentrated in outdoor scenes.

\noindent
\textbf{Temporal continuity and video duration}. 
From a temporal perspective, VideoCube divides 500 videos into time-continuous and time-intermittent. Canceling the continuous motion hypothesis breaks the temporal boundaries and extends the proportional timeline to a flexible one. For example, a 3-minute video of the SOT task can only record a 3-minute event. In contrast, a 3-minute video can be edited to reflect a story for more than an hour in the GIT task, increasing the richness of video content. As shown in Figure~\ref{fig:distribution} (c), video duration in VideoCube can be equally divided into four categories ranging from 3 minutes to 20 minutes, which is much higher than the existing video-based datasets.

\noindent
\textbf{Motion modes}. 
VideoCube records the two principal motion modes for each video. The 1000 motion modes are divided into 61 categories, as shown in Figure~\ref{fig:distribution} (d). 

Figure~\ref{fig:benchmark-compare} (e-f) shows the distribution of motion modes in VideoCube and LaSOT \cite{LaSOT}. Obviously, the total number of motion modes in VideoCube (61) is much larger than LaSOT (33). Besides, the statistical results of LaSOT are mainly concentrated in the most common modes, while the statistical results of VideoCube are distributed in a variety range. The long tail of distribution results in VideoCube indicates our work includes more rare movements. 

We believe that the 6D principle provides a scientific guide for the data collection, which increases the richness of video content and helps users quickly restore the narration from the six elements, improving the practicality of VideoCube. Figure~\ref{fig:example} illustrates the representative frames of this dataset.

\subsubsection{Annotation principle}

\begin{figure}[h!]
  \centering
  \includegraphics[width=0.95\linewidth]{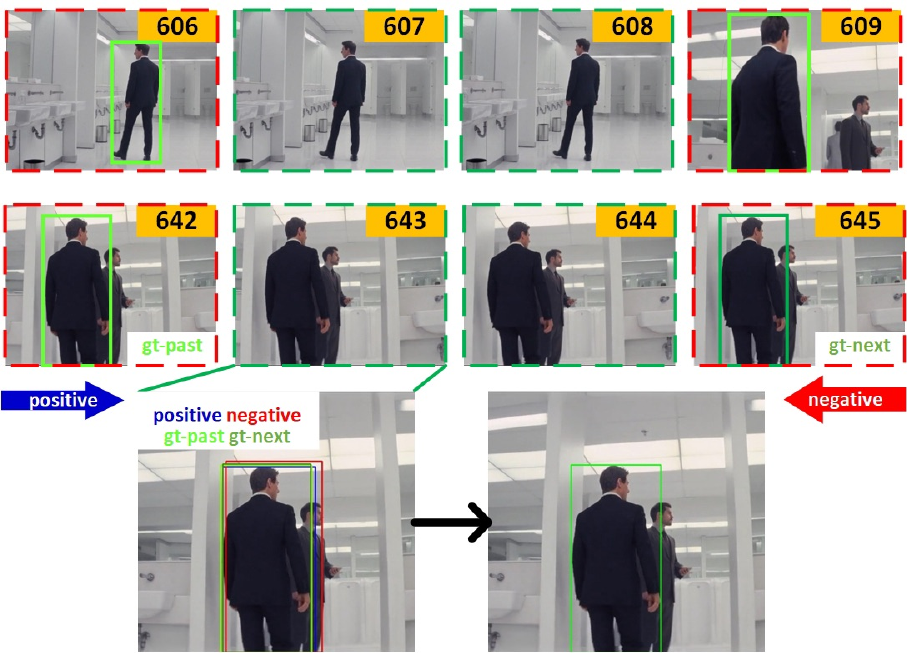}
  \caption{Examples of automatic annotation.}
  \label{fig:prdimp}
  \end{figure}

We use manual labeling and automatic algorithm for data annotation. The professional annotation team manually labels every three frames at a frequency of 10 Hz. After that, the PrDiMP \cite{PrDiMP} algorithm automatically provides labels for the rest two frames between manually labeled frames, as shown in Figure~\ref{fig:prdimp}.

\begin{figure}[h!]
  \centering
  \includegraphics[width=0.95\linewidth]{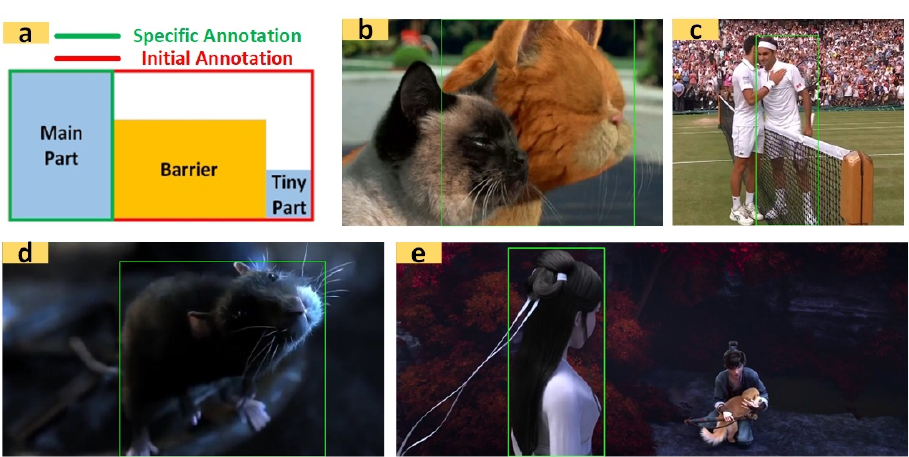}
  \caption{Examples of specific rules in VideoCube annotations.
  (\textbf{a}) Example of a tiny area.
  (\textbf{b})Garfield's transparent beard and a tiny part of the left side.
  (\textbf{c})Federer's right hand.
  (\textbf{d})The mouse's transparent beard and a slender tail.
  (\textbf{e})The white long ribbon.
  }
  \label{fig:special}
  \end{figure}

\begin{figure}[h!]
  \centering
  \includegraphics[width=0.95\linewidth]{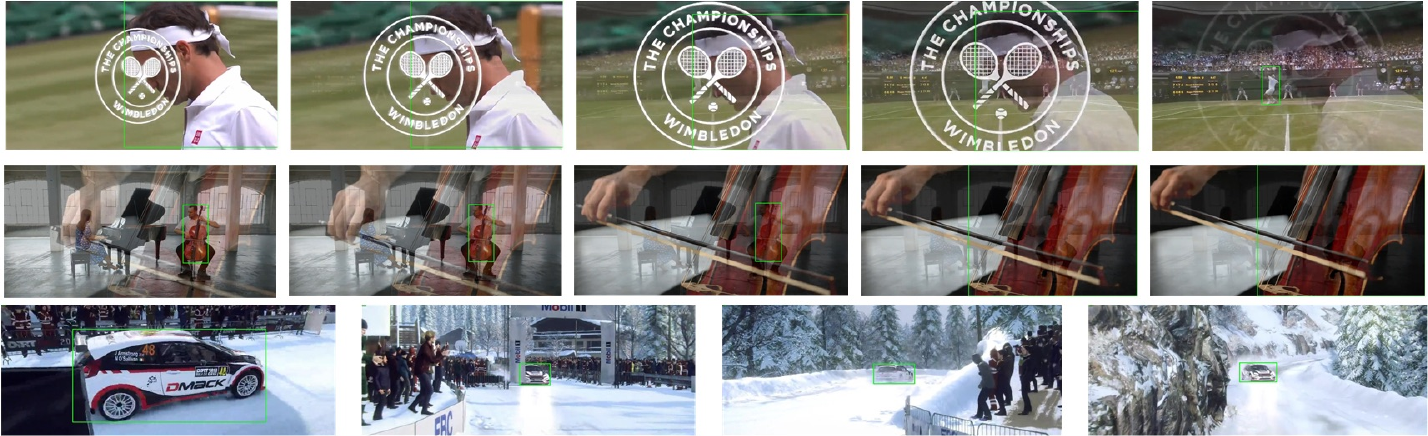}
  \caption{Examples of transitions in VideoCube annotations.
  (The first and second rows belong to the \emph{slow transition}, while every two frames of the third row is a \emph{fast transition})}
  \label{fig:transition}
  \end{figure}  

\noindent
\textbf{Manual annotation}. 
A professional project team rigorously labels VideoCube. The annotation process observes the following rules: (1) if the specific instance appears in the frame, the visible part of the instance is marked with the tightest bounding box; (2) if the instance is not in the frame, an absent label is marked.
Besides the two main rules above, some exceptions require individual labeling rules. We summarize the exceptional cases of high-frequency occurrences. The examples are provided in Figure~\ref{fig:special} (a): (1) \emph{Tiny area}: if the instance is divided into multiple areas by obstacles and labeling the tiny area will contain many obstacle pixels, the tiny part is discarded, and only labels the central area (Figure~\ref{fig:special} (a), (b) and (c)). (2) \emph{Transparent objects}: transparent beards of cats or mice are not marked (Figure~\ref{fig:special} (b) and (d)). (3) \emph{Slender and broad swinging objects}: the tail of a mouse or a long ribbon of a person are not marked (Figure~\ref{fig:special} (d) and (e)).

VideoCube also provides the instance absent label, the occlusion label, and the starting points of all shots. The transition is divided into two types: \emph{fast transition} and \emph{slow transition}. Transitions that occurred in two successive frames without motion continuity are considered as \emph{fast transition}, and the start frame of each new shot is labeled as a \emph{shot-cut}. Dissolve, fade-in, and fade-out between two scenes are \emph{slow transition}, and all frames belonging to the interim stage are labeled. Examples of transitions are shown in Figure~\ref{fig:transition}.

\begin{figure}[h!]
  \centering
  \includegraphics[width=0.95\linewidth]{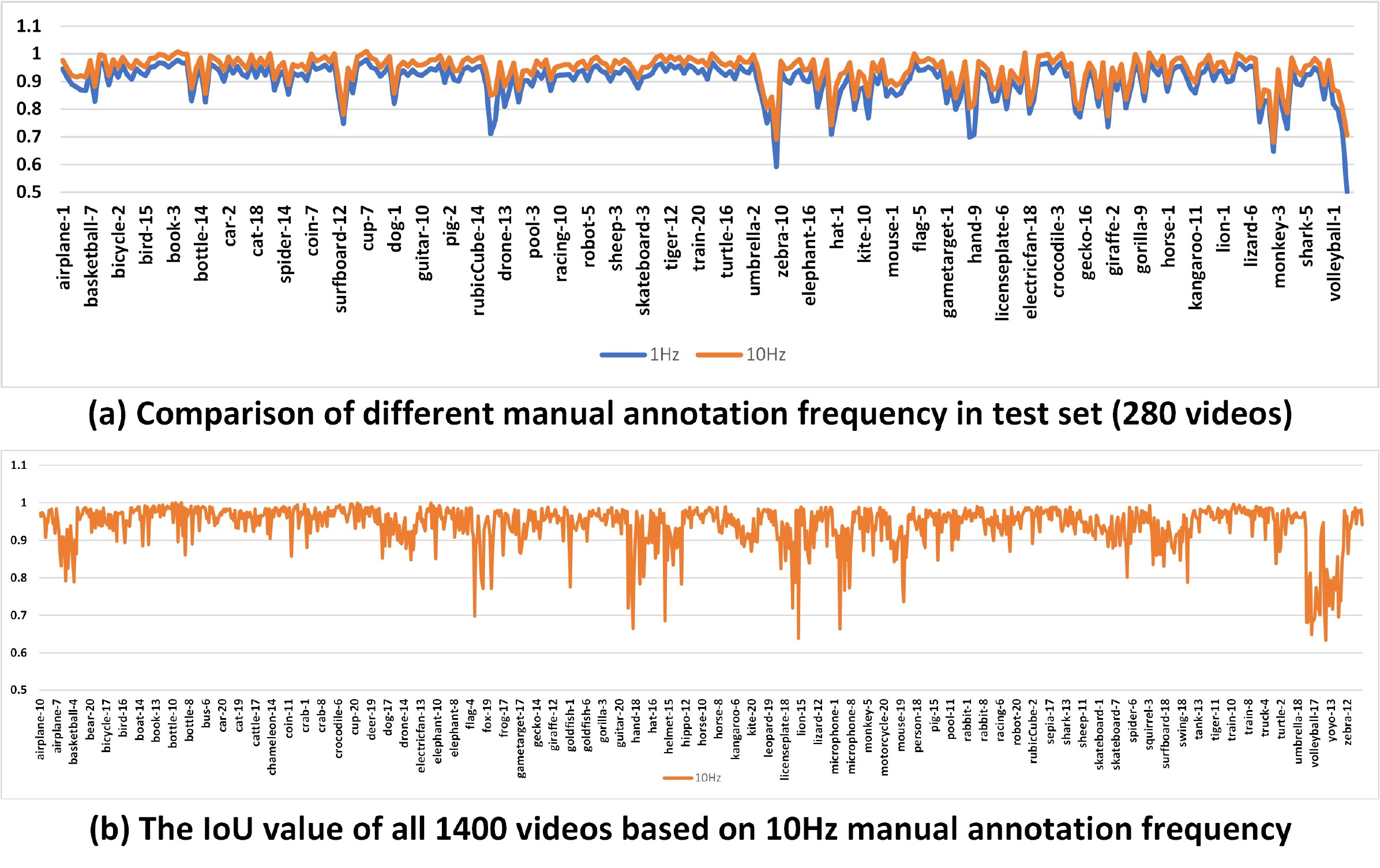}
  \caption{The experimental result of the automatic annotation on LaSOT. 
  }
  \label{fig:lasot}
  \end{figure} 

\noindent
\textbf{Automatic annotation}. 
The execution steps and completion strategies of the automatic annotation algorithm are shown in Figure~\ref{fig:prdimp}. The red dashed box represents a manually annotated frame, while the green dashed box represents an automatically completed frame. For the first row, the annotation team labels \#606 and \#609, and records the target position. Since the shot-switching occurs from \#608 to \#609, an extra transition tag is needed for \#609 to indicate the beginning of a new shot. The second row explicates the process of labeling \#643 via PrDiMP \cite{PrDiMP}. The target position of two nearest frames with manual annotation is marked as \emph{gt-past} (\#642) and \emph{gt-next} (\#645). In this sequence, PrDiMP \cite{PrDiMP} runs twice with forward order (from \#642 to \#645) and backward order (from \#645 to \#642) and records target location as \emph{positive} and \emph{negative}. We design several strategies to synthesize the position parameters of positive, negative, gt-past, and gt-next for different situations, then obtain the coordinate of instance in \#643. Algorithm 1 presents the framework for generating the automatic labels.

To verify the effect of the above strategy, we select LaSOT \cite{LaSOT} as the experiment dataset. It is a large-scale, long-term tracking dataset with a 30Hz manual label frequency (provide the manual label for each frame). The first version is released in 2019 with 1400 videos (total of 3.5M frames), while the new version in 2020 is expanded to 1550 videos (total of 3.87M frames). In this experiment, we select the first version to verify the performance of the automatic label method.
Figure~\ref{fig:lasot} presents the experimental result of the automatic annotation on LaSOT. The blue line in Figure~\ref{fig:lasot} (a) represents 1Hz manual annotation frequency with the automatically generated result for the middle 29 frames; the orange line represents 10Hz manual annotation frequency with the automatically generated result for the two middle frames. The average IoU score based on the 1Hz complementation plan is 0.9, while the average IoU score based on the 10Hz is 0.95. Figure~\ref{fig:lasot} (b) shows the IoU value of all 1400 videos based on 10Hz manual annotation frequency with the automatically generated result for the two middle frames. 
It indicates that the 1Hz manual labeling frequency is too sparse to provide a factual basis of the automatic completion scheme (such as TrackingNet) or evaluation (such as OxUvA). The 10Hz manual labeling frequency with a suitable automatic annotation mechanism can improve efficiency and provide a human-level annotation via an effective algorithm.

\begin{algorithm}[t!]
  \footnotesize
  \caption{Framework of generate the automatic annotation}
  \SetKwInput{KwInput}{Input}                % Set the Input
  \SetKwInput{KwOutput}{Output}              % set the Output
  \DontPrintSemicolon
    \KwInput{$P_{gt}$: previous mannually labeled bounding-box; 
    $N_{gt}$: next mannually labeled bounding-box; 
    $B_{pos}$: bounding-box generated in forward order; 
    $B_{neg}$: bounding-box generated in backward order
    }
  
    \KwOutput{$B$: bounding-box of present frame}
  
    calculate $D_{1} = DIoU(P_{gt}, N_{gt})$
  
    calculate $D_{2} = DIoU(B_{pos}, B_{neg})$
  
    \tcc{Situation 1: a high value of $D_{1}$ indicates miniature movement, and the location can be directly calculated}
    \If{$D_{1} \geq \tau{_{1}}$} 
      {
          $B = average(P_{gt},N_{gt})$ 
          return $B$
      }
  
      calculate $E_{1} = Enclose(P_{gt}, N_{gt})$ 
  
    \tcc{Situation 2: a high value of $D_{2}$ indicates normal movement, and this is the most common situation. We assume that the motion range of instance in intermediate frame does not exceed $E_{1}$}
    \If{$D_{2} \geq \tau{_{2}}$}
    {
          \If {both $B_{pos}$ and $B_{neg}$ are enclosed by $E_{1}$}
          {$B = average(B_{pos}, B_{neg})$}
          \ElseIf {$B_{pos}$ or $B_{neg}$ is enclosed by $E_{1}$}
          {$B = B_{pos}$ or $B = B_{neg}$}
          \ElseIf {both $B_{pos}$ and $B_{neg}$ are outside $E_{1}$}
          {$B = average(P_{gt},N_{gt})$}
          return $B$
    }
    \tcc{Situation 3: the situation does not belong to the above two conditions indicates rapid movement or shot-switching}
  
    \If {presnet frame is the last two frame in a shot}
    {$B = average(B_{pos},P_{gt})$}
  
    \ElseIf {presnet frame is the first two frame in a shot}
    {$B = average(B_{neg},N_{gt})$}
  
    \Else
    {
        calculate $D_{3} = DIoU(P_{gt}, B_{pos})$ 
  
        calculate $D_{4} = DIoU(N_{gt}, B_{neg})$
  
        \If {$D_{3} \geq D_{4}$}
        {$B = Intersection(E_{1}, B_{pos})$} 
        \Else 
        {$B = Intersection(E_{1}, B_{neg})$}
    }
    return $B$
  \end{algorithm}

\subsubsection{Checkout flow} 

\begin{figure}[h!]
  \centering
  \includegraphics[width=0.95\linewidth]{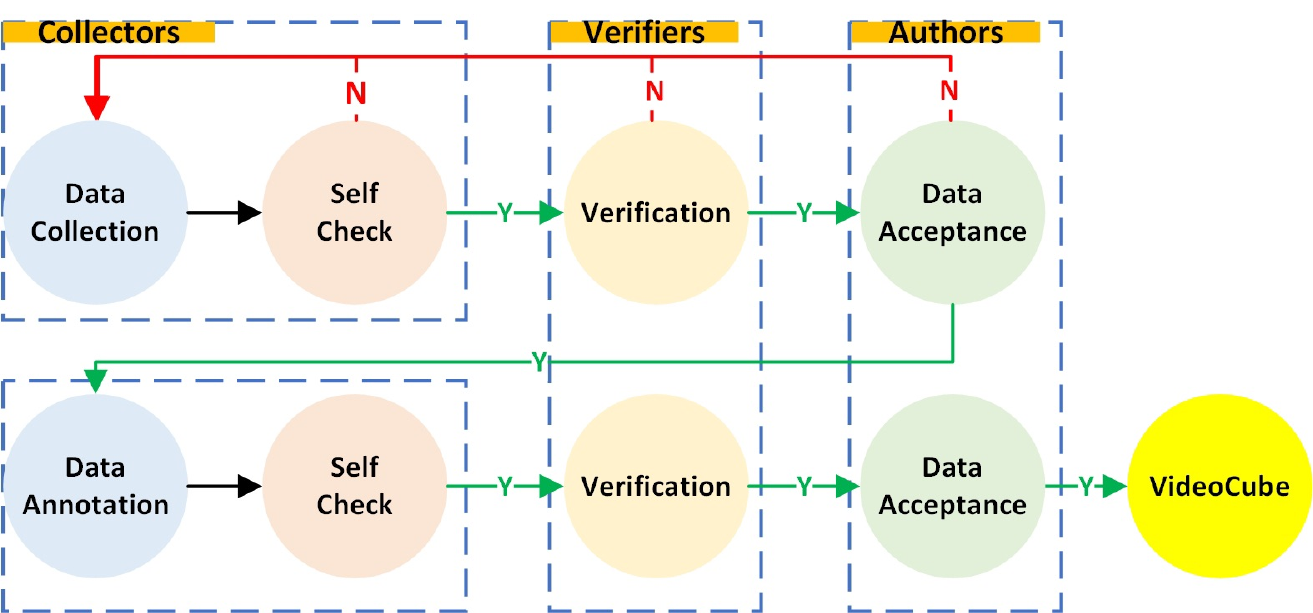}
  \caption{The framework of data checkout process. }
  \label{fig:checkout}
  \end{figure}  

We implement a strict data review process to ensure the quality of the benchmark. The construction process is divided into two tasks: data collection and data annotation. Professional collectors and annotators are trained to comprehend the GIT task's characteristics and complete the preliminary work with a self-inspection process. The verifiers review the submitted data as the second-round verification. Finally, the authors judge whether to accept or reject it in the third-round confirmation. As shown in Figure~\ref{fig:checkout}, any rejection in self-check, verification, or data acceptance will result in the re-collection. We believe the three-round verification mechanism can generate a high-quality dataset.

\subsubsection{Attribute selection}

Twelve attributes are selected in VideoCube to enable further performance analysis: 

\noindent
\textbf{Instance Absent (IA)}, the instance is out-of-view or fully occluded by other objects, manually labeled by annotators. 

\noindent
\textbf{Shot-cut (SC)}, frame belongs to slow transition or fast transition, as the starting point for a new shot and is manually labeled by annotators. 

\noindent
\textbf{Instance Occlusion (IO)}, more than 10$\%$ of the instance is occluded, manually labeled by annotators. 

\noindent
\textbf{Illumination Variation (IV)}, illumination changes between previous and current frames. We use the Shade of Gray algorithm \cite{ShadeofGray} of color constancy to calculate the correction matrix $C_i$ between the current illumination and the standard illumination. Multiplying the original frame $F_i$ and the correction matrix $C_i$ can obtain the frame $S_i$ under standard illumination. The gamma correction factor is 2.2 and the power is 6 in the correction matrix calculation. Subsequently, the cosine similarity between the vectors $C_i$ and $O_i = [1,1,1]$ is calculated as the illumination standard $i_i$ of the current frame: $i_i = \frac{C_i\bullet O_i}{\parallel C_i\parallel\times\parallel O_i\parallel}$. 
$i_i$ is a quantization value of \textbf{Illumination Estimation (IE)}, and $i_i<0.99$ means special illumination in current frame.The difference in absolute value between previous frame $i_{i-1}$ and current frame $i_i$ is $\Delta i_i$, $\Delta i_i > 0.0001$ means illumination variation in continuous frames. 

\noindent
\textbf{Blur Variation (BV)}, quantization of the sharpness variation between previous and current frames. The variation of the Laplacian \cite{Laplacian} is selected to calculate the blur degree. We first convert the current image $F_i$ into a grayscale image $G_i$, then convolve $G_i$ with a specific Laplacian kernel $L$, and calculate the variance of the response result $v_i$ —
this value is used as an index of sharpness.
Images with $v_i < 100$ can be considered blurry; otherwise are clear. Besides, the difference in absolute value between $v_{i-1}$ and $v_i$ is $\Delta v_i$, while $\Delta v_i > 1.5$ means blur variation.

\noindent
\textbf{Scale Variation (SV)}, indicator for measuring changes in instance scales. The size of instance in current frame is $s_{i}=\sqrt{w_{i} h_{i}}$, and $s_{i} \notin [50,750]$ will be considered as \textbf{Special Scale (SS)}. $\Delta s_i$ is calculated by the difference of absolute value between $s_{i-1}$ and $s_i$, and $\Delta s_i > 30$ signifies scale variation.

\noindent
\textbf{Ratio Variation (RV)}, indicator for characterizing the target deformation and rotation. The aspect ratio of instance in current frame is $r_{i}=\frac{h_i}{w_i}$, and $r_{i} \notin [\frac{1}{3},3]$ will be considered as \textbf{Special Ratio (SR)}. Same as the previous calculation process, $\Delta r_i > 0.2$ stands for ratio variation.

\noindent
\textbf{Fast Motion (FM)}, an index $d_{i}=\frac{\left\|c_{i}-c_{i-1}\right\|_{2}}{\sqrt{s_{i} s_{i-1}}}$ aims to measure the instance motion speed. The motion of object in current frame is $d_i$, where the $c_i$ indicates the center of bounding box. Since $d_i$ has reflected the dynamic relationship between $F_i$ and $F_{i-1}$, it can be used to reflect the motion variation between two frames directly. $d_i>0.2$ will be treated as fast motion.

\noindent
\textbf{Correlation Coefficient (CC)}, a metric used to measure the similarity between $F_i$ and $F_{i-1}$. In this paper, we use the Pearson product-moment correlation coefficient(PPMCC) $p_i = \rho_{i, i-1}=\frac{\operatorname{cov}(F_i, F_{i-1})}{\sigma_{F_i} \sigma_{F_{i-1}}}$. The numerator calculates the covariance of the current frame $F_i$ and the previous frame $F_{i-1}$, and the denominator is the product of the standard deviation. The correlation coefficient reflects the changes between consecutive frames and has been normalized; it can be used as an attribute index directly. $p_i > 0.8$ signifies the correlation between the continuous two frames is strong.

Twelve attributes can be divided into three types: filtering attributes, self attributes, and dynamic attributes. Instance absent (IA) and shot-cut (SC) are filtering attributes to remove frames that are unsuitable for metrics calculation in experiments. Instance occlusion (IO), illumination estimation (IE), special scale (SS), and special ratio (SR) are self attributes that only reflect the characteristics of the current frame rather than embody the dynamic variations. Blur variation (BV), illumination variation (IV), scale variation (SV), ratio variation (RV), fast motion (FM), and coefficient of correlation (CC) are dynamic attributes that contain the dynamic relationship between consecutive frames.

\subsection{Evaluation system}

\begin{figure}[h!]
  \centering
  \includegraphics[width=0.95\linewidth]{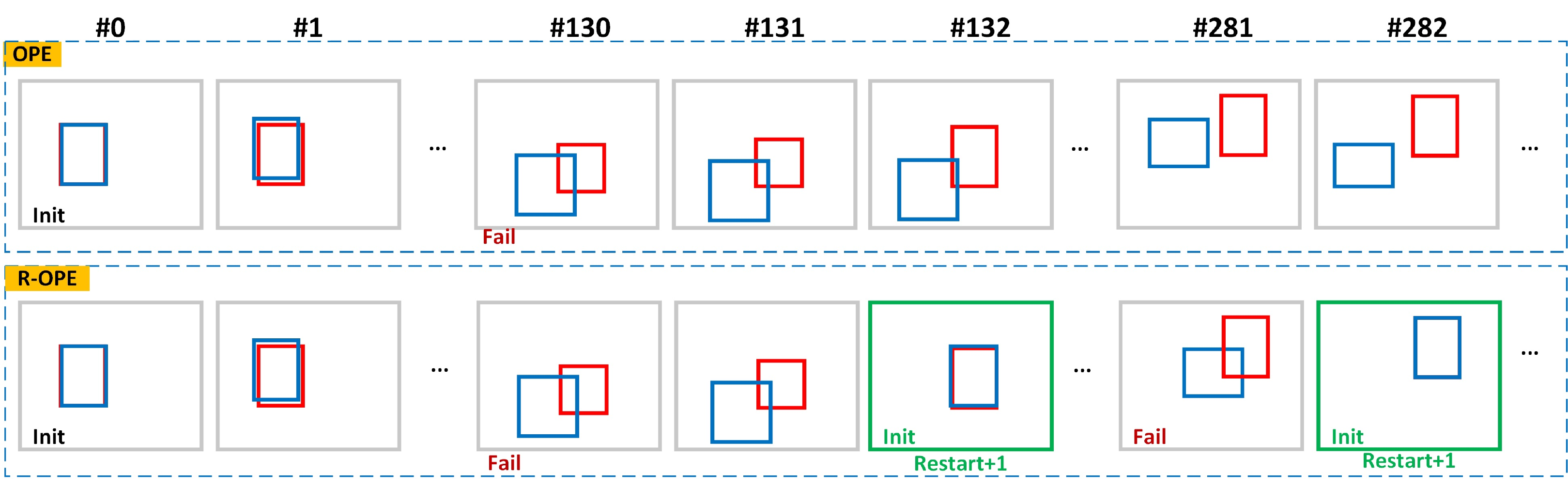}
  \caption{Comparison of the two evaluation mechanisms. The first row illustrates traditional OPE mechanism, and the second row illustrates R-OPE mechanism with failure detection and tracker restart.
  }
  \label{fig:restart}
  \end{figure}

The following two sections introduce the evaluation system and performance measurement of the GIT task. The evaluation system aims to judge the model's capabilities (such as accuracy and robustness) through a reasonable evaluation method. The performance measurement focuses on quantitatively mapping the model capabilities through scientific calculation to accomplish more in-depth analysis and sort the results via numerical values. 

\subsubsection{One-pass evaluation (OPE)}
The evaluation protocol of OTB \cite{OTB2015} benchmark has six categories: three normal processes, including one-pass evaluation (OPE), temporal robustness evaluation (TRE), spatial robustness evaluation (SRE), and three restart processes involving one-pass evaluation with restart (OPER), temporal robustness evaluation with restart (TRER), spatial robustness evaluation with restart (SRER). Among them, the OPE method is defined as using the ground-truth in the first frame to initialize the model and continuously locate the target in subsequent frames. Subsequent tracking-based visual tasks (i.e., short-term and long-term tracking) are only distinguished in assignment settings but maintain the OPE method as the evaluation system. 

Numerous benchmarks listed in Table~\ref{table:benchmarks} except two short-term tracking benchmarks (OTB and VOT) only retain the OPE mode but discard the restart mechanism. However, the restart proposed by OTB does not perform real-time supervision but generates the OPER results based on a series of existing experimental results generated by the TRE method. Although VOT can complete a fail-detection in the algorithm running process, the re-initialization is performed directly without any design for selecting the restart frame.

\subsubsection{One-pass evaluation with restart mechanism (R-OPE)}

The restart mechanism is essential in evaluating the GIT task for the following reasons: (1) Videos in VideoCube have a longer average frame and include multiple challenging characteristics like shot-switching and scene-transferring. Thus, models are prone to fail in locating instances and cannot be reinitialized. (2) The count of restarts can be quantified as an indicator to measure the algorithm's robustness. A similar restart mechanism has been studied on monkeys by neuroscientists \cite{li2008unsupervised,cox2005breaking}. They replace the target $P$ with interference $N$ during the rapid eye movements of monkeys. After several repetitions, the observation of activities in the temporal cortex of monkeys indicates that the monkey has confused $P$ and $N$. Some online update algorithms continue learning the apparent characteristics of the instance during the tracking process. However, challenges like lens-switching mean the algorithm needs to expand the search range to relocate the instance (like rapid eye movement). Re-location may misidentify the interference as an instance and update the wrong sample. This situation is caused by the wrong instance updating rather than weak learning ability. Therefore, VideoCube includes two evaluation systems: traditional OPE and \emph{OPE with restart mechanism (R-OPE)}.

The foundation of the R-OPE mechanism is selecting restart frames. The selection process follows two principles: (1) The restart frame is manually annotated rather than automatically generated to ensure the label quality. (2) Rich instance features are contained in the restart frame to provide enough information for re-initialization.  We select the YOLACT++ \cite{bolya2019yolact} algorithm to segment each manually labeled bounding box, delete the background, match the remaining instance with the clear-cut query in the first frame, and finally determine frames with matching points exceeding a certain threshold as restart frames. According to statistics, 17.2\% of frames in VideoCube satisfy the filter conditions. 

Example of R-OPE mechanism is shown in Figure~\ref{fig:restart}. 
The first row illustrates the traditional OPE mechanism. The tracker is initialized in the first frame, in which the algorithm result (blue) and ground-truth (red) coincide. In the following tracking process, the IoU value of the algorithm result (blue) and ground-truth (red) is less than a threshold (usually 0.5) in the \#130, which indicates a failure. Since the OPE mechanism does not detect failure, the continued failure causes subsequent frames to be wasted. 
The second row is the R-OPE mechanism with failure detection and tracker restart. The green frame indicates an appropriate restart point. After the tracking failure is detected at \#130, the algorithm will be re-initialized at the nearest restart point (\#132), and subsequent sequences will continue to participate in the evaluation. When the tracking failure occurs in \#281, the algorithm will be restarted at \#282.

\begin{figure}[t!]
  \centering
  \includegraphics[width=0.95\linewidth]{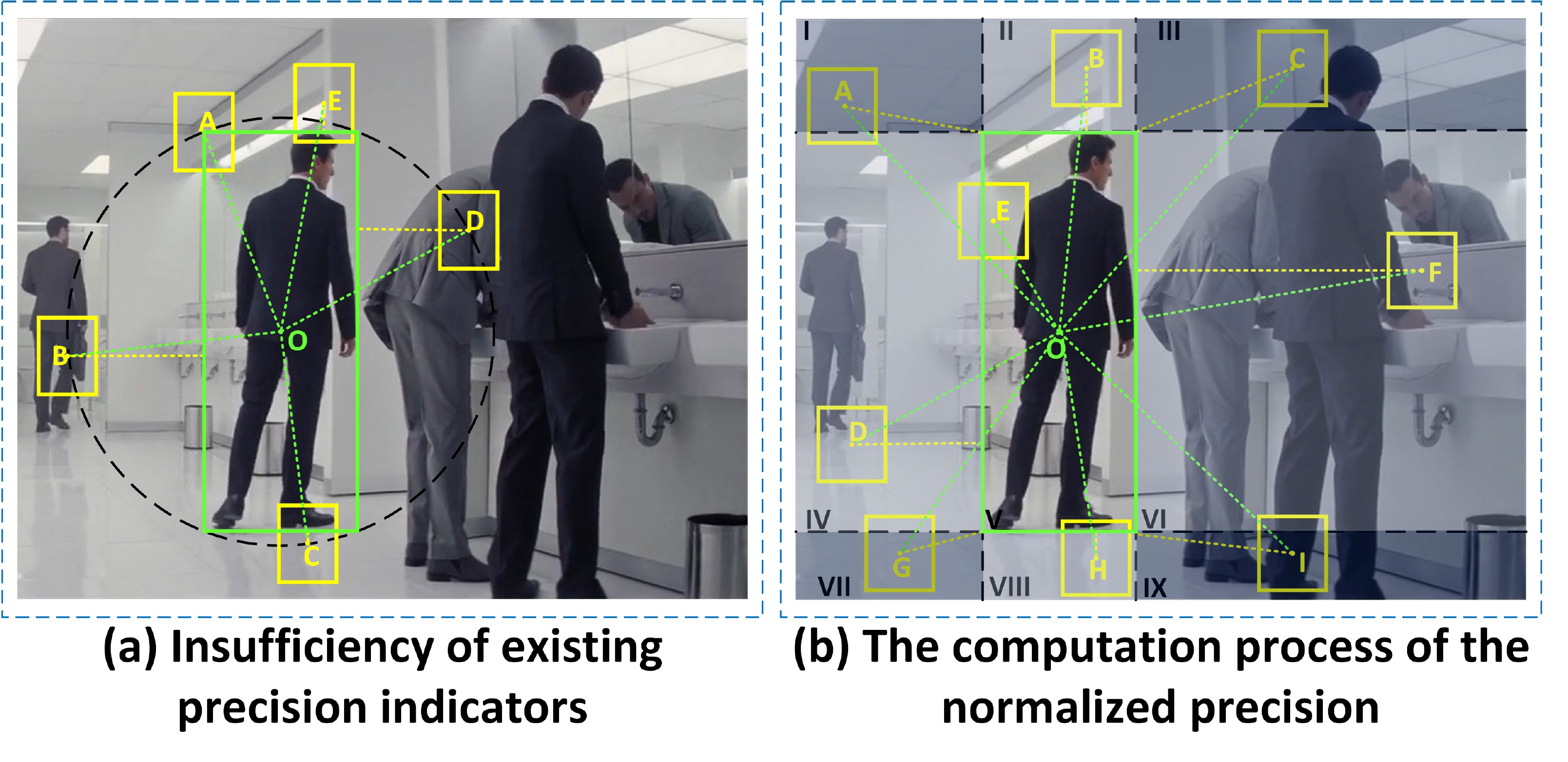}
  \caption{The counterexample of traditional precision metrics and the computation process of the normalized precision (N-PRE). 
  (\textbf{a}) Insufficiency of existing precision indicators. Common sense infers that algorithm $A$ has the highest accuracy while $B$ performs worst. However, the traditional precision (TRE) indicator results consider $A$, $B$, $C$, $D$ have the same precision and are better than $E$.
  (\textbf{b}) The computation process of the normalized precision (N-PRE). The ground-truth bounding box divides the screen into nine areas (\textbf{I} to \textbf{IX}). 
  Point $E$ falls into area \textbf{V} (ground-truth); the distance between $E$ and the $O$ point is considered the original precision value of tracker $E$. 
  For other trackers that fall into eight external areas, the original precision value is the sum of two parts. The first part is the distance between the center point and $O$ (shown as the green dashed line); the second part is the penalty item calculated by the shortest distance between the center point and the edge of the ground-truth box (shown as the yellow dashed line). To exclude the influence of instance size and frame resolution, we select the maximum value of all screen points to normalize the result.
  }
  \label{fig:normprec}
  \end{figure}

\subsection{Evaluation metrics}

Similar to the metrics used by most SOT benchmarks \cite{LaSOT,OTB2015,GOT-10k,OxUvA}, we first utilize the precision plot and the success plot to measure the performance of the algorithms for OPE and R-OPE mechanisms. 

\subsubsection{Precision plot}

Tradition precision (PRE) measures the center distance between the predicted result $p^t$ and the ground-truth $b^t$ in pixels. Calculating the proportion of frames whose distance is less than the specified threshold and drawing the statistical results based on different thresholds into a curve generates the precision plot. Typically, trackers are ranked on 20 pixels \cite{OTB2015, LaSOT}. 
However, the object scale is influenced by target size and image resolution but ignored by the original PRE score. Thus, two new benchmarks \cite{LaSOT, TrackingNet} adopt the ground-truth scale (width and height) to normalize the center distance. Specifically, the height difference and width difference between two center points are divided by the ground-truth shape before calculating the distance. This operation solves the target scale influence on PRE calculation but is still not comprehensive enough.

Figure~\ref{fig:normprec} (a) presents a counterexample. The green rectangle represents the ground-truth, where point $O$ denotes the center point. Assume that five yellow rectangular boxes show the prediction results of the five algorithms. To eliminate the influence of other factors, here assumes the prediction results have only position differences. $OA$, $OB$, $OC$, and $OD$ are the same, while $OE$ is slightly larger. 
The precision scores of tracker $A$, $B$, $C$, $D$ based on two existing metrics (directly using the center point distance or only using the current ground-truth size for normalization) are the same, while tracker $E$ is worse.
Nevertheless, from Figure~\ref{fig:normprec} (a), we can directly judge that $A$ and $E$ perform better than $B$. The calculation results are contrary to common sense because the target aspect ratio affects accuracy but is ignored by existing metrics. For non-square bounding boxes, only the center point distance cannot quantify the tracking accuracy accurately.

To deal with the above problem, we propose a new precision metric N-PRE. Explicitly, we select the center distance as the original precision if the tracker center point falls into the ground-truth rectangle. Algorithms with a predicted center outside the ground-truth rectangle will also calculate the shortest distance between its center and the ground-truth edge.
As shown in Figure~\ref{fig:normprec} (b), the original precision value of tracker $E$ is $OE$, while other trackers are calculated by two parts (center distance represented by the green dashed line, and the penalty item represented by the yellow dashed line). 
Subsequently, we quantify the original precision value to the $[0,1]$ interval; 0 represents the tracker center point is $O$, while 1 represents the score of the farthest point in the current frame (upper right point).
In Figure~\ref{fig:normprec} (a), the performance of tracker $A$ evaluated via N-PRE is the best while tracker $B$ is the worst. It is consistent with reality.

\subsubsection{Success plot}
To get the success rate (SR), we first calculate the Intersection over Union (IoU) of the predicted result $p^t$ and the ground-truth $b^t$. Frames with an overlap rate greater than a specified threshold are defined as successful tracking, and the SR measures the percentage of successfully tracked frames under different overlap thresholds. The statistical results based on different thresholds create the success plot. Besides, we implement two more success scores based on Generalized IoU (GIoU \cite{GIoU}) and Distance IoU (DIoU \cite{DIoU}), aiming to provide a comprehensive scientific evaluation.

\subsubsection{Robustness}
For the R-OPE mechanism, we propose a new evaluation indicator to evaluate robustness. Specifically, we define robustness as $R=\frac{1}{N}\sum_{i=1}^{N}[S(\frac{1}{ {\rho}_i } )(1-\frac{I_i}{R_i})]$. $N$ represents the number of videos participating in the evaluation, ${\rho}_i$ indicates the correlation coefficient of the i-th video, $R_i$ means the total number of restart frames selected for this video, and  $I_i$ denotes the number of restarts of the tracker.

\begin{figure*}[h!]
  \centering
  \includegraphics[width=0.7\textwidth]{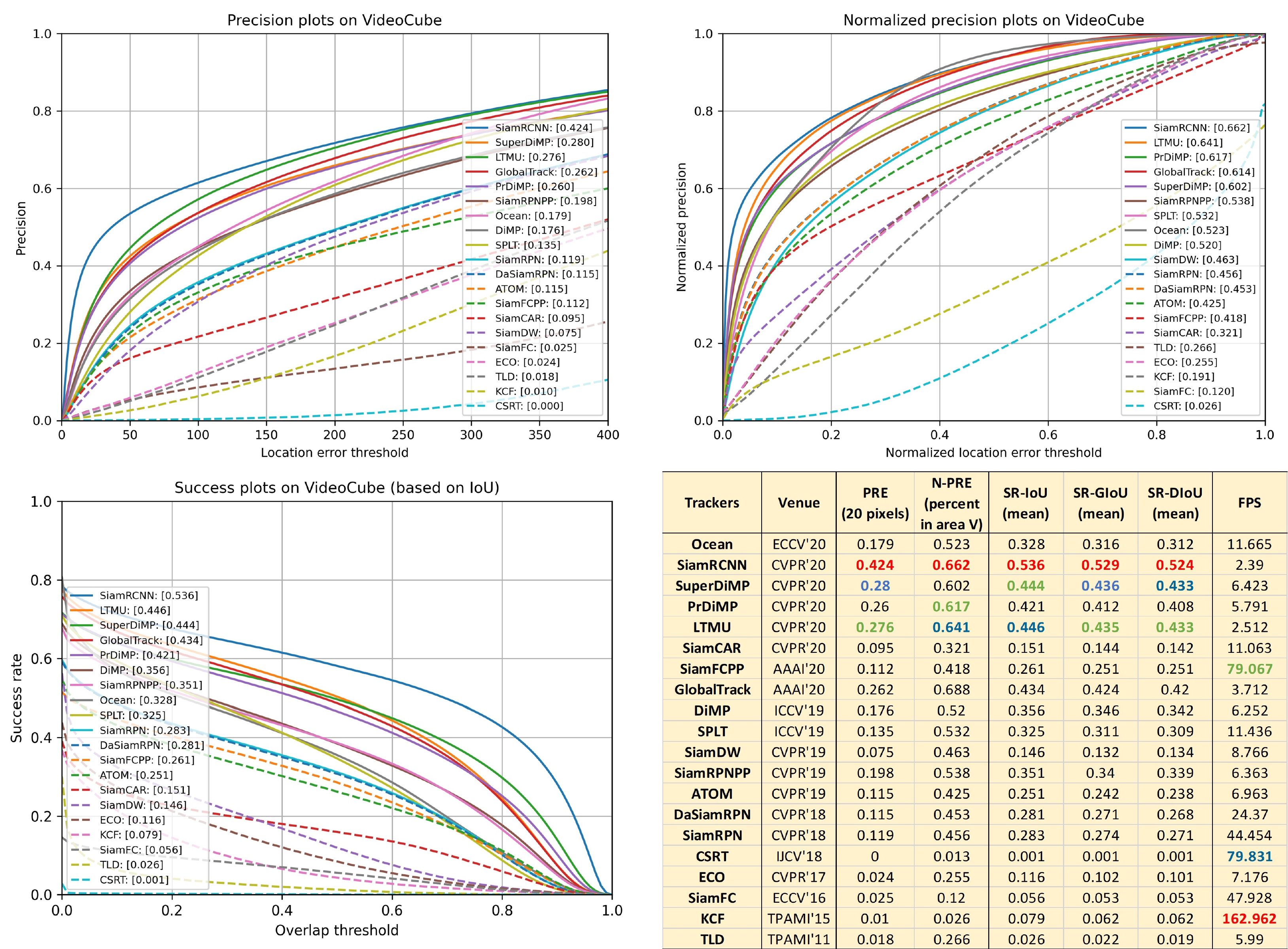}
  \caption{Standard experiments in OPE mechanism, evaluated by precision (PRE) plot, N-PRE plot, and success plot. The red, blue, and green in the tables represent the first, second, and third placed algorithms of each indicator.
  }
  \label{fig:ope}
  \end{figure*}

\begin{figure*}[h!]
  \centering
  \includegraphics[width=0.7\textwidth]{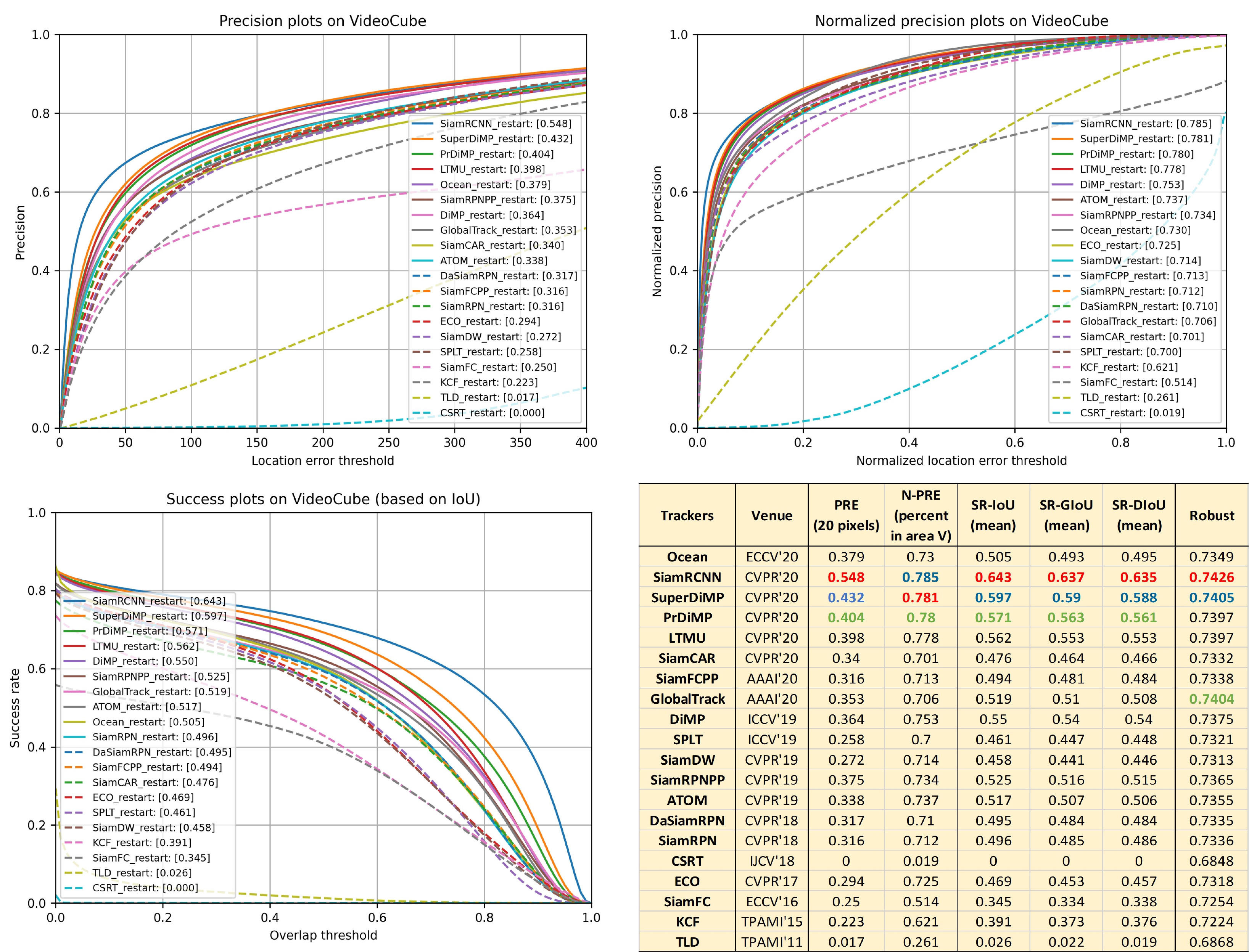}
  \caption{Standard experiments in R-OPE mechanism, evaluated by precision (PRE) plot, N-PRE plot, and success plot. The red, blue, and green in the tables represent the first, second, and third placed algorithms of each indicator.
  }
  \label{fig:rope}
  \end{figure*}

\section{Experiments}
\label{sec:experiments}

\begin{figure*}[t!]
  \centering
  \includegraphics[width=0.90\textwidth]{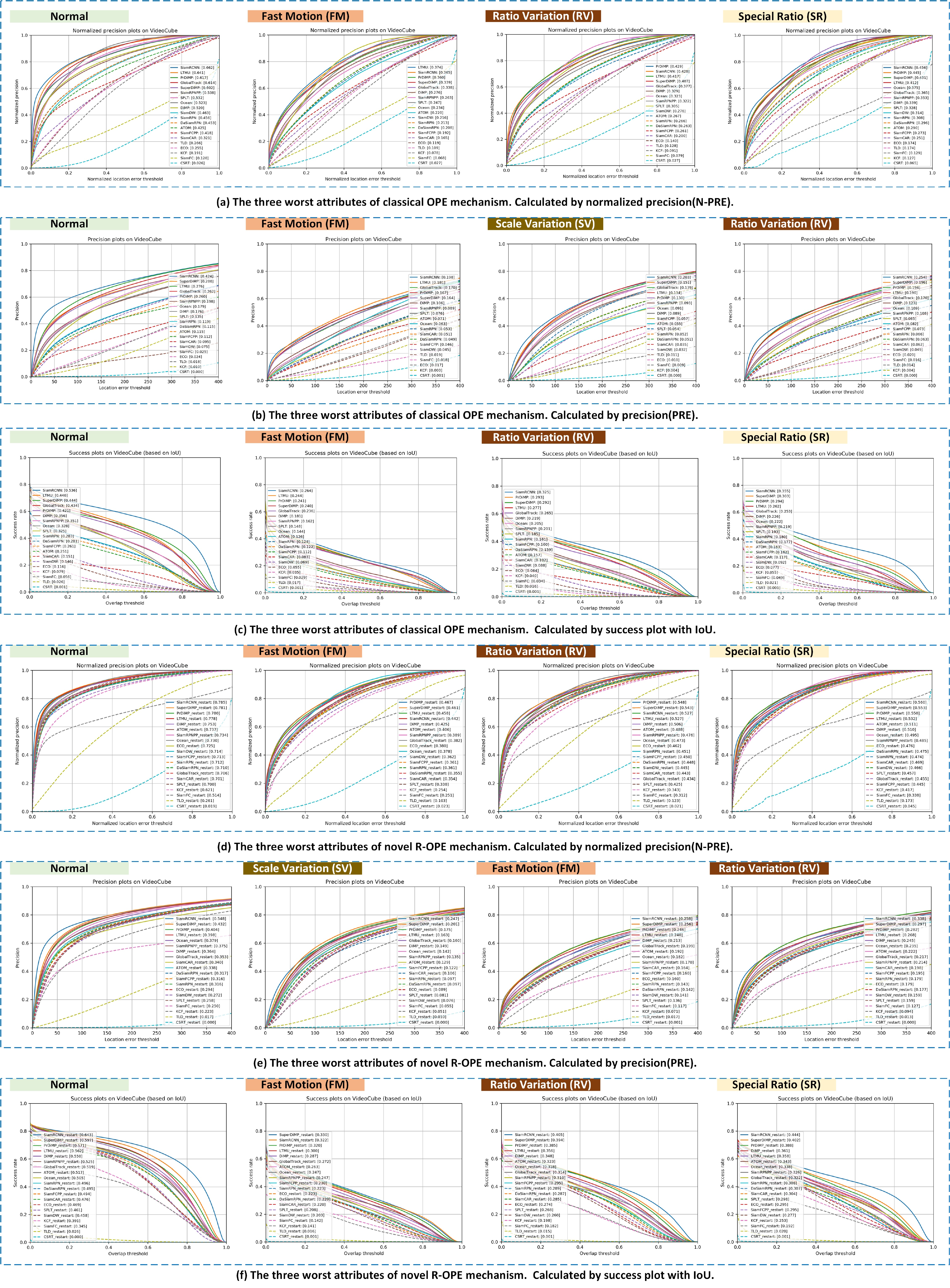}
  \caption{The attribute performance. (\textbf{a}) to (\textbf{c}) illustrates the performance of the three worst attributes in classical OPE mechanism by different evaluation metrics. (\textbf{d}) to (\textbf{f}) illustrates the performance of the three worst attributes in R-OPE mechanism by different evaluation metrics. }
  \label{fig:attribute-result}
  \end{figure*}

\begin{figure*}[t!]
  \centering
  \includegraphics[width=0.85\textwidth]{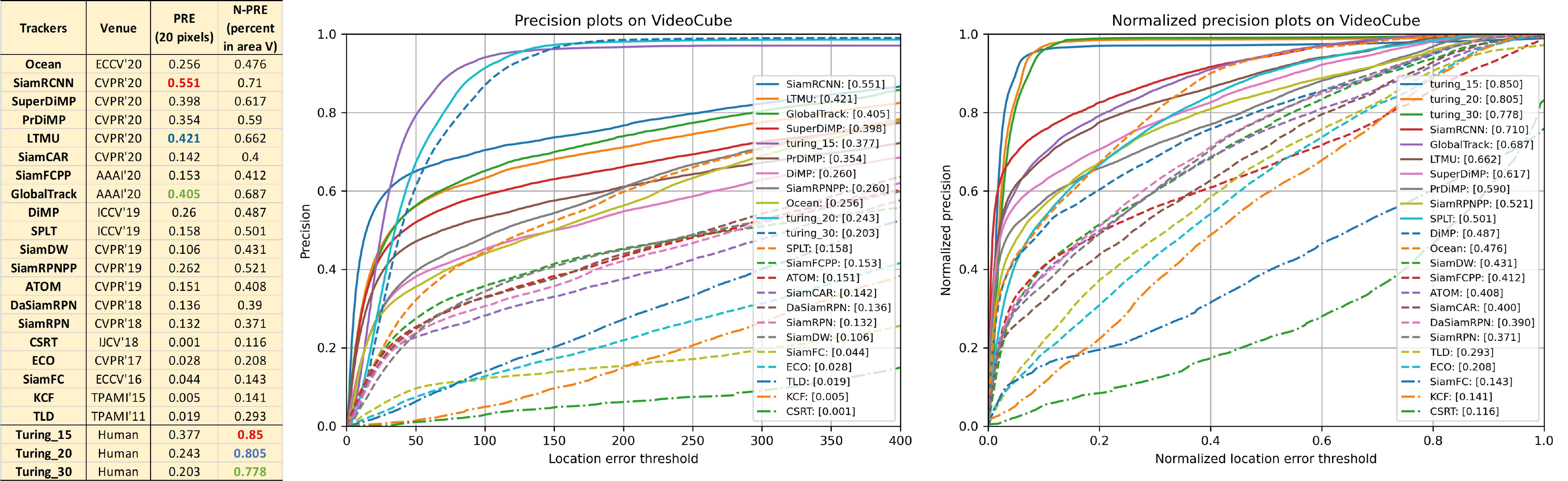}
  \caption{Eye-tracking experiments in OPE mechanism, evaluated by precision (PRE) plot and N-PRE plot. The red, blue, and green in the tables represent the first, second, and third placed algorithms of each indicator.
  }
  \label{fig:eyeresult}
  \end{figure*}

We accomplish extensive experiments in this section and divide them into two parts:

\noindent
\textbf{Standard experiments}. 
We select 20 algorithms (Ocean \cite{Ocean}, SiamRCNN \cite{SiamRCNN}, SuperDiMP \cite{PrDiMP}, LTMU \cite{LTMU}, PrDiMP \cite{PrDiMP}, SiamCAR \cite{SiamCAR}, SiamFC++ \cite{SiamFC++}, SiamDW \cite{SiamDW}, GlobalTrack \cite{GlobalTrack}, DiMP \cite{DiMP}, SPLT \cite{SPLT}, SiamRPN++ \cite{SiamRPN++}, ATOM \cite{ATOM}, DaSiamRPN \cite{DaSiamRPN}, SiamRPN \cite{SiamRPN}, ECO \cite{ECO}, SiamFC \cite{SiamFC}, TLD \cite{TLD}, CSRT \cite{CSRT}, KCF \cite{KCF}) as baselines and conduct experiments under the OPE and R-OPE mechanisms. All algorithms are fully evaluated under two mechanisms to generate the precision plot and success plot.

\noindent
\textbf{Eye tracking experiments}. 
We apply an eye tracker machine to record and quantify the human visual tracking ability. The intelligence level of trackers can be measured by comparing human capacity with algorithm tracking results.

\subsection{Standard experiments}

Twenty trackers are selected as baseline models and evaluated on VideoCube. Given that most algorithms do not determine the instance absent, we first remove the frames that exclude the tracked instance. Besides, frames in the transition stage may include superimposed instances. To ensure the accuracy of the evaluation, we remove the transition frames as well.

\subsubsection{Overall performance}

Figure~\ref{fig:ope} and Figure~\ref{fig:rope} present the overall performance of trackers in OPE and R-OPE mechanisms. The scores and rankings of algorithms under these two mechanisms are pretty distinct, confirming that the two evaluation mechanisms' focuses are different. 
For evaluation results in OPE (Figure~\ref{fig:ope}), the algorithm scores are low since the VideoCube allows lens switching and scene transferring, causing the jump change of the target position in consecutive frames. Most algorithms strongly depend on continuous motion assumption and usually use local search to locate the target, thus performing worse when the position variation occurs.
The R-OPE mechanism restarts algorithms at the next restart point after detecting the failure (Figure~\ref{fig:rope}). Its precision plot and success plot focus on evaluating the local-search ability, while the robustness score obtained via quantifying the number of restarts reflects the global-search ability.

\subsubsection{Attribute performance}

VideoCube selects twelve attributes to describe the challenges in the GIT task and divides them into three categories: filtering attributes, self attributes, and dynamic attributes. We provide twelve attribute labels for each frame to fully capture the difficulty factors. The detailed results are demonstrated in Figure~\ref{fig:attribute-result}. It is clear that compared with other attributes, \textit{fast motion (FM)}, \textit{ratio variation (RV)}, \textit{special ratio (SR)}, and \textit{scale variation (SV)} challenge the performance of trackers.

\subsection{Eye tracking experiments}

\begin{figure}[h!]
  \centering
  \includegraphics[width=0.95\linewidth]{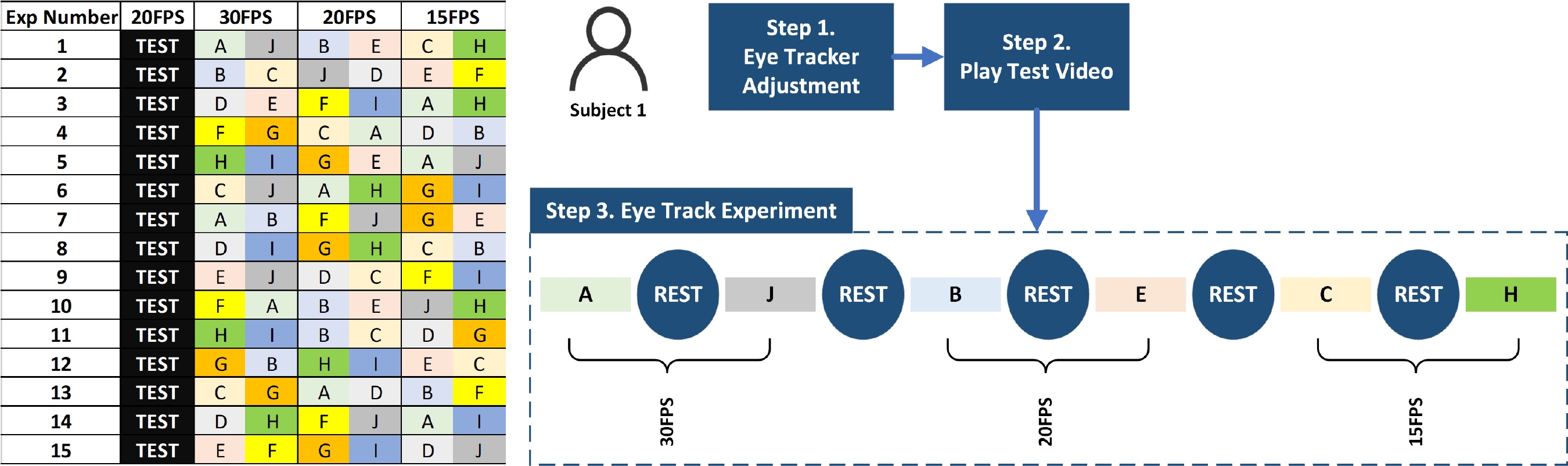}
  \caption{The process of eye-tracking experiment.
  Ten videos (A-J) with different difficulty, duration, instance types, space classes, motion modes are played to the subject at three speeds (15FPS, 20FPS, and 30FPS). Fifteen subjects track the test videos at three rates. 
  }
  \label{fig:eye-process}
  \end{figure}

\begin{figure}[h!]
  \centering
  \includegraphics[width=0.95\linewidth]{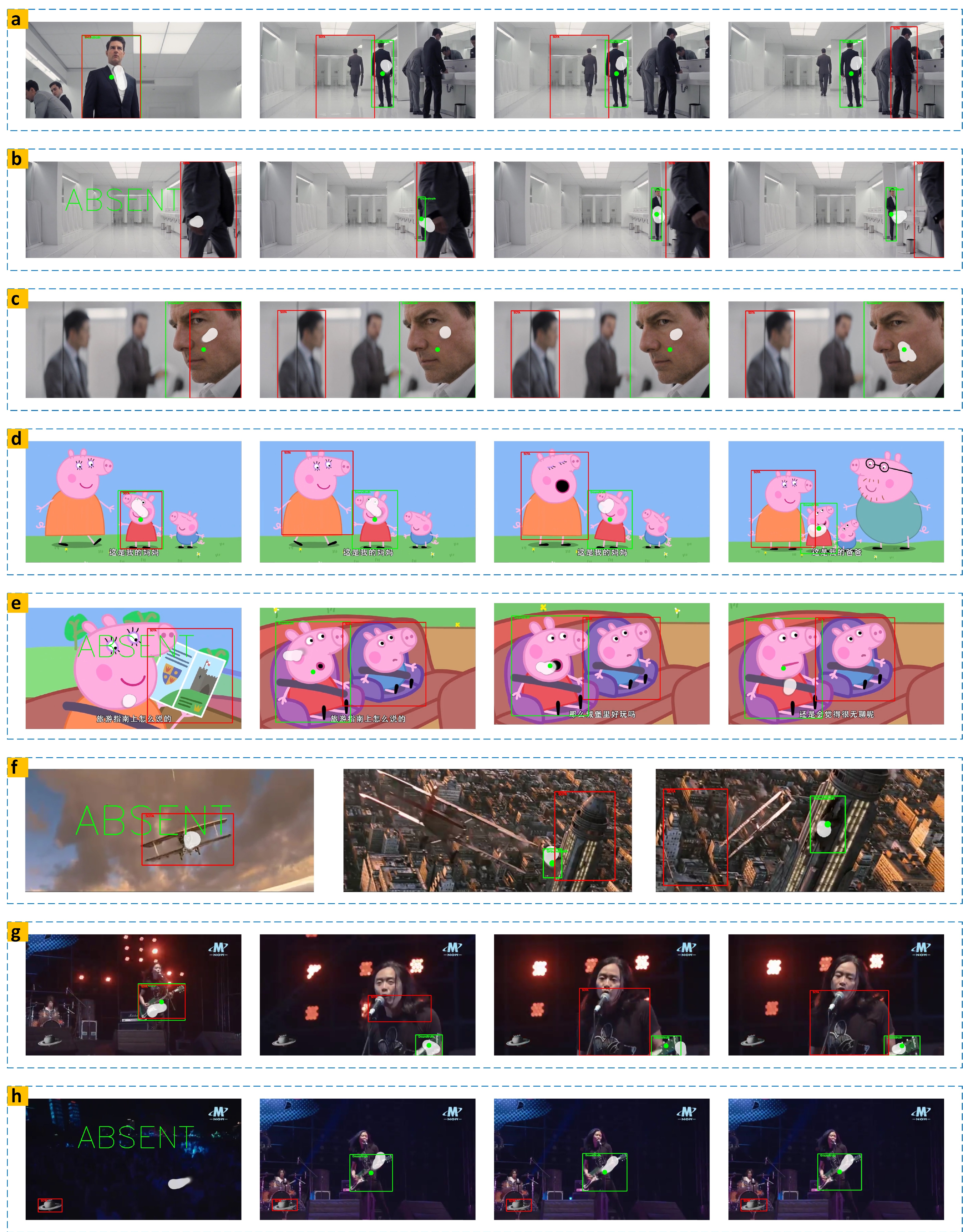}
  \caption{Some examples of human visual tracking ability better than SOTA algorithm.
  (\textbf{a}, \textbf{e}, and \textbf{f}) When the \textit{transition} occurs, the human can locate the target immediately, while the algorithm drifts to a similar object.
  (\textbf{b}) When the \textit{obstruction} is removed, the human can locate the target immediately, but the algorithm keeps tracking the obstruction.
  (\textbf{c} and \textbf{d}) When \textit{same-category objects} appear on the screen, people can distinguish between the target and similar items, but the algorithm fails.
  (\textbf{g} and \textbf{h}) The emergence of \textit{transitions} and the \textit{complex illumination environment} makes the algorithm unable to locate the target, but humans can quickly identify the target (guitar) through auxiliary information (the guitarist).
  }
  \label{fig:human-better}
  \end{figure}
  
\begin{figure}[h!]
  \centering
  \includegraphics[width=0.95\linewidth]{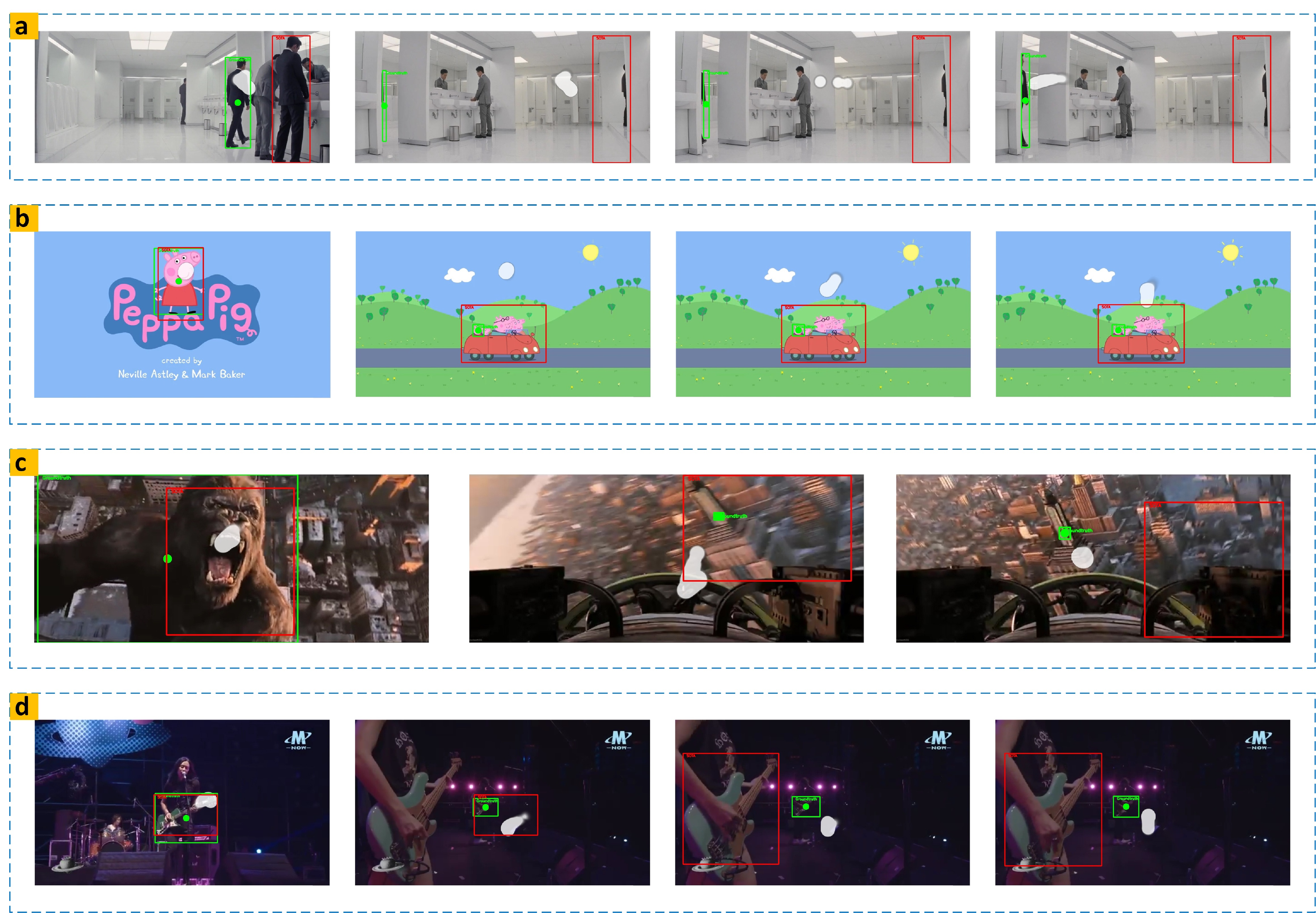}
  \caption{Some examples where both the human and the SOTA algorithm fail.
  (\textbf{a}) The \textit{transition} causes the jumps of target position, and the \textit{occlusion} factor in the new scene makes it impossible for both humans and algorithms to locate the target quickly.
  (\textbf{b}) The \textit{transition}, the presence of \textit{interference} and the \textit{tiny size} make it challenging to locate the target.
  (\textbf{c}) The \textit{transition}, the frame \textit{blur} due to \textit{fast movement} and the \textit{tiny size} make it challenging to locate the target.
  (\textbf{d}) The \textit{transition}, the \textit{tiny size} and the \textit{complex illumination environment} make it challenging to locate the target.
  }
  \label{fig:both-worse}
  \end{figure}

Unlike traditional visual tracking experiments that only evaluate algorithms with performance rather than intelligence, we design an eye-tracking experiment to judge human visual tracking ability and measure machine intelligence via comparison.

Ten videos with different difficulty, duration, instance types, space classes, motion modes are played to the subject at three speeds (15FPS, 20FPS, and 30FPS). 
We select 15 human subjects to track the test video at three rates. 
We have obtained the approvals of all the human participants. Every participant has signed an informed consent form before the experiment.
Firstly, each subject should calibrate the eye tracker machine to ensure that the instrument can accurately detect the sightline. Secondly, the test video appears in the screen center, and the subject should focus on the target in the first frame, then press the play button. After that, the subject needs to concentrate on the target and maintain tracking accuracy. The subject has a rest time to relieve visual fatigue between two videos.
The eye tracker machine records the eye movement of subjects, and the focus of sight is used to calculate the precision score and generate precision plots. 

Figure~\ref{fig:eye-process} illustrates the detailed process of eye-tracking experiments, which consists of three steps. (1) The subject calibrates the eye tracker machine (Tobii Eye Tracker) to ensure that the instrument can accurately detect the sightline. (2) A TEST video appears in the screen center. The subject should focus on the target in the first frame, press the play button, and concentrate on maintaining tracking accuracy. TEST video aims to help subjects familiarize themselves with the test process. (3)  The subject begins the formal experiment by tracking six different videos. A break between two videos is needed to ensure the effectiveness of the experiment.

Figure~\ref{fig:eyeresult} presents the precision plots of humans and 20 trackers in OPE mechanisms.Turing\_15, Turing\_20, and Turing\_30 represent human scores at 15FPS, 20FPS, and 30FPS, respectively. 
We can draw the following conclusions through comparison:
(1) The calculation methods and sequencing principles of traditional precision (PRE) scores have multiple problems. PRE measures the center distance between the predicted result and the ground-truth in pixels, but ignores the impact of target size and video resolution (for detailed analysis, please refer to the methods chapter). This makes the ranking threshold with 20 pixels unreasonable. In the precision plot of Figure~\ref{fig:eyeresult}, human performance is far lower than algorithms, contrary to our common sense. Since the deviation of the eye tracker machine may exceed 20 pixels in several situations, 20 pixels are too strict by comparing with the image resolution and target size of videos in VideoCube.
(2) The normalized precision plot shows that the human visual tracking ability is worse than tracking algorithms for strict precision requirements. The reason may be the deviation of the eye tracker machine and the human attention (for person target, subjects prone to focus on the head instead of the torso). When the accuracy requirements are moderately reduced, the human visual ability will quickly exceed algorithms and remain stable.

Besides, Figure~\ref{fig:human-better} explains how the human eye outperforms the SOTA algorithms. Humans can quickly locate the target in challenging factors such as transitions, similar objects, occlusion, and complex illumination occur, while the algorithms always fail or drift to similar instances. The experiment presents that algorithms need to enhance the robustness in challenging environments to achieve human-like tracking.
Figur~\ref{fig:both-worse} presents cases where human eyes and algorithms fail, indicating that some extreme environments challenge both humans and algorithms to accomplish target tracking.

\section{Conclusion}
\label{sec:conclusion}
To help trackers locate the target more like humans, we analyze the fundamentals of measuring the intelligence level and summarize the limitations of existing benchmarks. In this paper, we (1) propose the GIT task to explicitly model the human visual tracking ability, (2) build the VideoCube benchmark to create a challenging experimental environment close to the real world, and (3) finally design a scientific evaluation procedure to measure the tracking performance of humans and machines.
The experimental results show that there is still a definite gap between trackers and humans. Still, we believe the general online platform treats human tracking capabilities as a baseline to evaluate the machine intelligence level, guiding the research to accomplish human-like trackers in the future.

\ifCLASSOPTIONcompsoc
  \section*{Acknowledgments}
\else
  \section*{Acknowledgment}
\fi

This work is supported in part by the National Natural Science Foundation of China (Grant No. 61721004 and No.61876181), the Projects of Chinese Academy of Science (Grant No. QYZDB-SSW-JSC006), the Strategic Priority Research Program of Chinese Academy of Sciences (Grant No. XDA27000000) and the Youth Innovation Promotion Association CAS.

\bibliography{GIT}
\bibliographystyle{IEEEtran}

\begin{IEEEbiography}[{\includegraphics[width=1in,height=1.25in,clip,keepaspectratio]{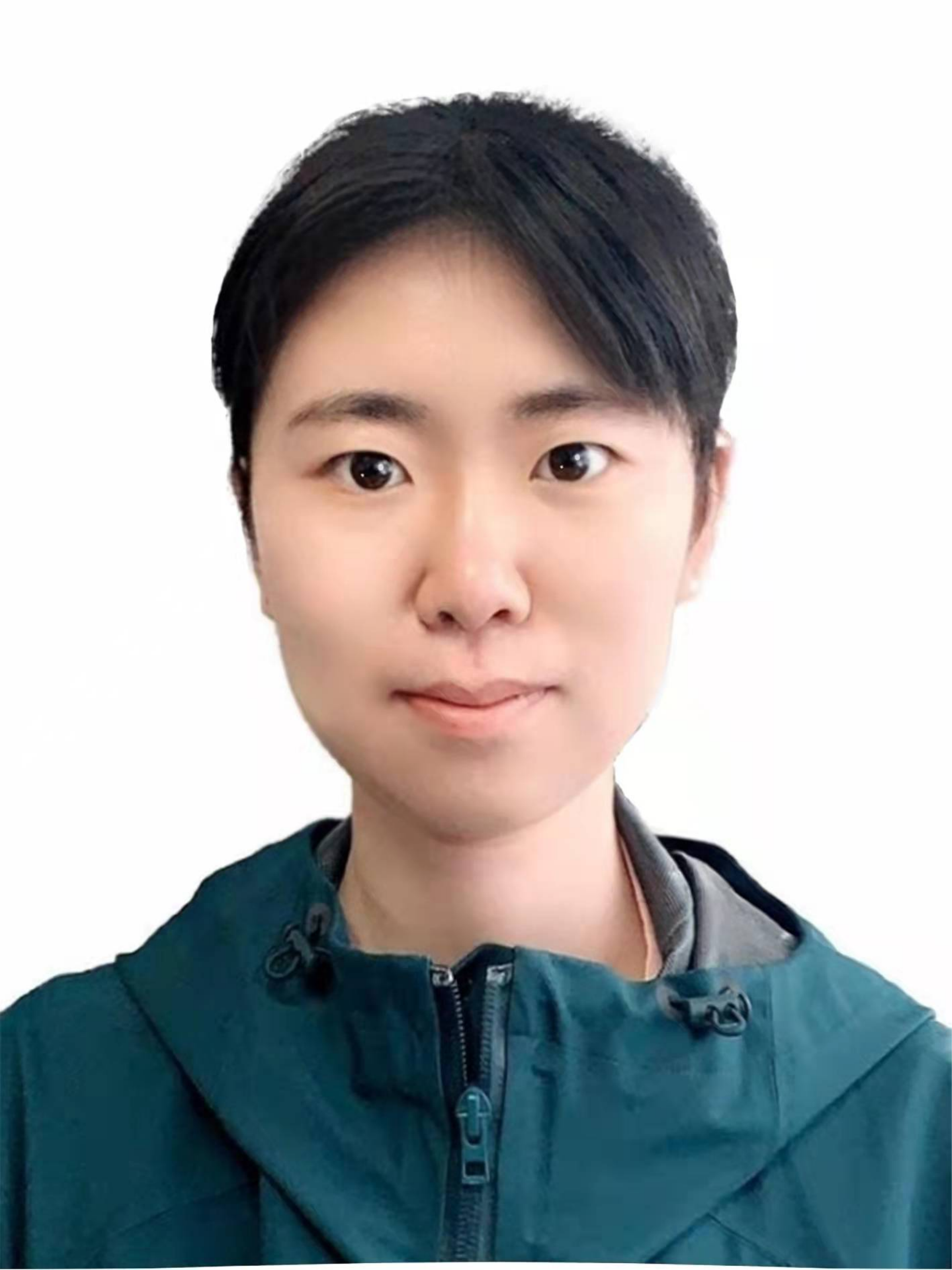}}]{Shiyu Hu}
  received her BSc from Beijing Institute of Technology (BIT) and MSc from the University of Hong Kong (HKU). In September 2019, she joined the Institute of Automation of the Chinese Academy of Sciences (CASIA), where she is currently studying for her doctorate. Her current research interests include pattern recognition, computer vision, and machine learning.
\end{IEEEbiography} 

\begin{IEEEbiography}[{\includegraphics[width=1in,height=1.25in,clip,keepaspectratio]{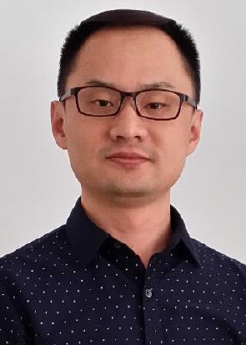}}]{Xin Zhao}
  received the Ph.D. degree from the University of Science and Technology of China. He is currently an Associate Professor in the Institute of Automation, Chinese Academy of Sciences (CASIA). His current research interests include pattern recognition, computer vision, and machine learning. He received the International Association of Pattern Recognition Best Student Paper Award at ACPR 2011. He received the 2nd place entry of COCO Panoptic Challenge at ECCV 2018.
  \end{IEEEbiography} 

\begin{IEEEbiography}[{\includegraphics[width=1in,height=1.25in,clip,keepaspectratio]{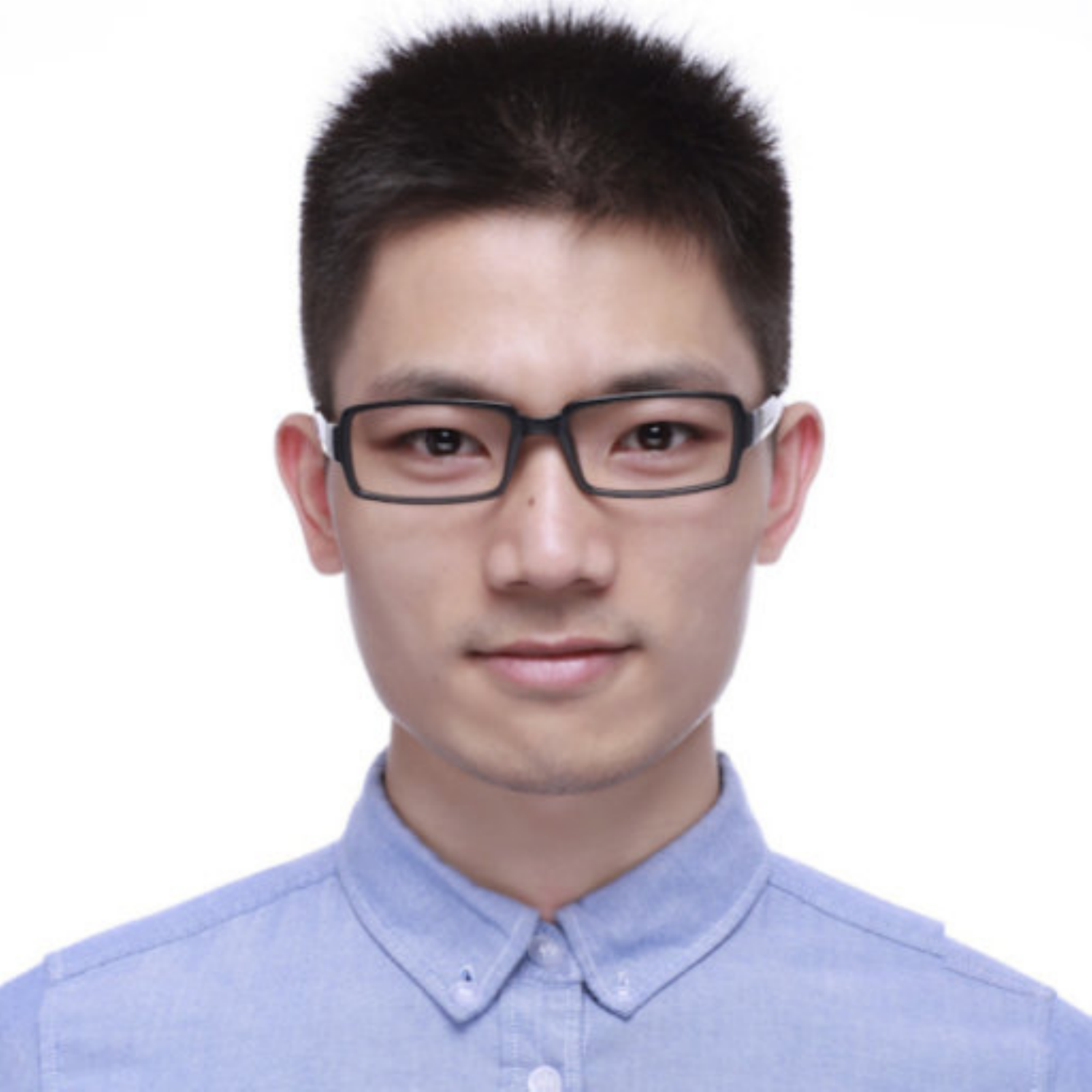}}]{Lianghua Huang}
received the Ph.D. degree from the Institute of Automation, Chinese Academy of Sciences (CASIA). He has published 17 papers in the areas of computer vision and pattern recognition at international journals and conferences including TPAMI, CVPR, ICCV, AAAI, ACMMM, TMM, TIP, TCYB, and ICME. He received the 1st place entry of VizWiz VQA challenge at CVPR 2021. His current research interests include representation learning and generative modeling in computer vision.
\end{IEEEbiography}

\begin{IEEEbiography}[{\includegraphics[width=1in,height=1.25in,clip,keepaspectratio]{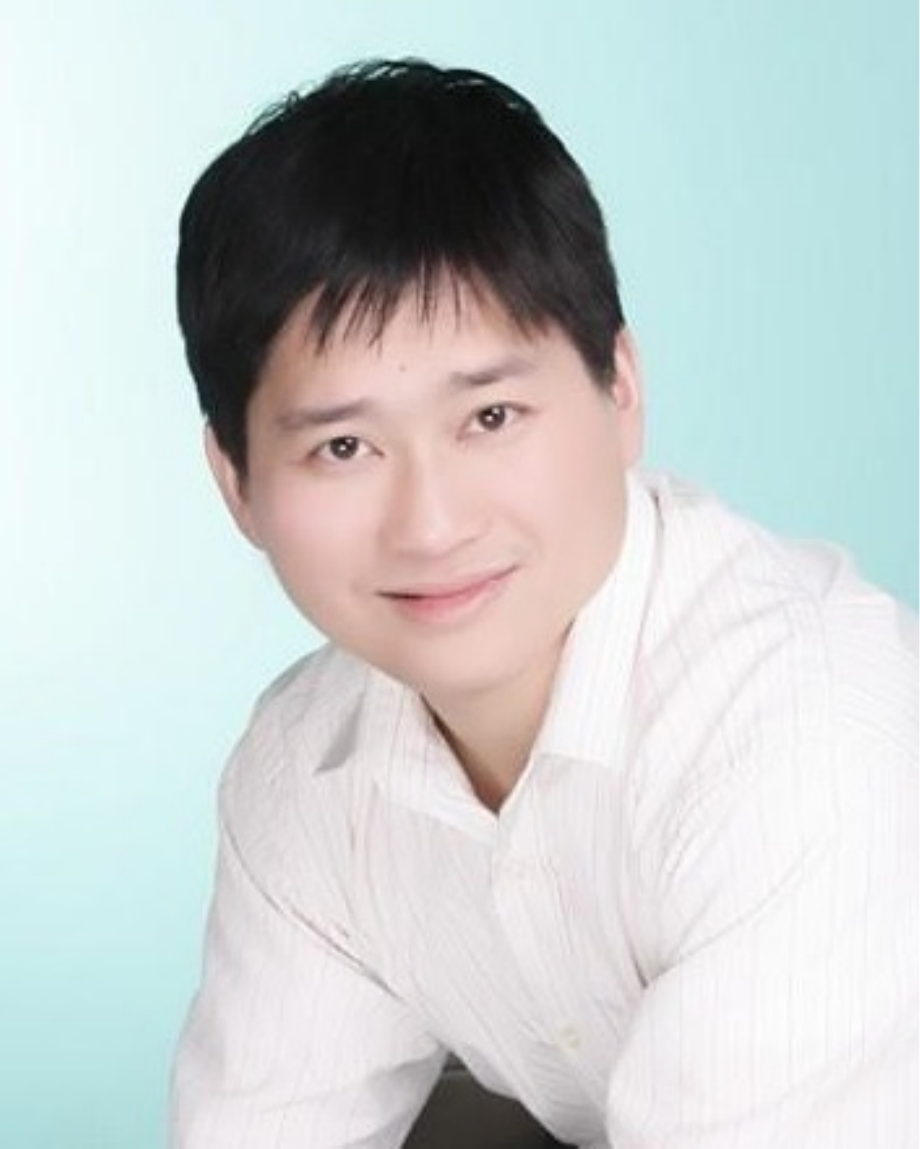}}]{Kaiqi Huang}
received the B.Sc. and M.Sc. degrees from the Nanjing University of Science Technology, China, and the Ph.D. degree from Southeast University. He is a Full Professor in the Center for Research on Intelligent System and Engineering (CRISE), Institute of Automation, Chinese Academy of Sciences (CASIA). He is also with the University of Chinese Academy of Sciences (UCAS), and the CAS Center for Excellence in Brain Science and Intelligence Technology. He has published over 210 papers in the important international journals and conferences, such as the IEEE TPAMI, T-IP, T-SMCB, TCSVT, Pattern Recognition, CVIU, ICCV, ECCV, CVPR, ICIP, and ICPR. His current researches focus on computer vision,pattern recognition and game theory, including object recognition, video analysis, and visual surveillance. He serves as co-chairs and program committee members over 40 international conferences, such as ICCV, CVPR, ECCV, and the IEEE workshops on visual surveillance. He is an Associate Editor of the IEEE TRANSACTIONS ON SYSTEMS, MAN, AND CYBERNETICS: SYSTEMS and Pattern Recognition.
\end{IEEEbiography} 

\end{document}